\pdfoutput=1
\documentclass[11pt]{article}

\usepackage[final]{acl}

\usepackage{times}
\usepackage{latexsym}

\usepackage[T1]{fontenc}

\usepackage[utf8]{inputenc}

\usepackage{microtype}

\usepackage{inconsolata}

\usepackage{graphicx}

\usepackage{lipsum} 
\usepackage{subcaption}
\usepackage{tikz}
\usepackage{bm}
\usepackage{amsmath,amssymb,amsthm}
\usepackage{booktabs}

\usepackage{tcolorbox}
\usepackage{lipsum} 

\newtheorem{definition}{Definition}

\usepackage[framemethod=TikZ]{mdframed}
\mdfdefinestyle{Takeaway}{
  linecolor=black,
  outerlinewidth=1pt,
  roundcorner=5pt,
  skipabove=5pt,
  innertopmargin=\dimexpr9pt-\topskip\relax,
  innerrightmargin=20pt,
  innerleftmargin=20pt,
  backgroundcolor=gray!20!white}
\mdtheorem[style=Takeaway]{takeaway}{Takeaway}
\newcommand{\take}[1]{\begin{takeaway}#1\end{takeaway}}

\title{What Did I Do Wrong? \\ Quantifying LLMs' Sensitivity and Consistency to Prompt Engineering}

\author{Federico Errica \and Giuseppe Siracusano \and Davide Sanvito \and Roberto Bifulco \\
        NEC Italia \and NEC Laboratories Europe \\ \texttt{name.surname@neclab.eu}}

\begin{document}
\maketitle
\begin{abstract}
Large Language Models (LLMs) changed the way we design and interact with software systems. Their ability to process and extract information from text has drastically improved productivity in a number of routine tasks. Developers that want to include these models in their software stack, however, face a dreadful challenge: debugging LLMs' inconsistent behavior across minor variations of the prompt. We therefore introduce two metrics for classification tasks, namely \textit{sensitivity} and \textit{consistency}, which are complementary to task performance. First, sensitivity measures changes of predictions across rephrasings of the prompt, and does not require access to ground truth labels. Instead, consistency measures how predictions vary across rephrasings for elements of the same class. We perform an empirical comparison of these metrics on text classification tasks, using them as guideline for understanding failure modes of the LLM. Our hope is that sensitivity and consistency will be helpful to guide prompt engineering and obtain LLMs that balance robustness with performance. 
\end{abstract}

\section{Introduction}
\label{sec:introduction}

\begin{quote}
\textit{There are only two hard things in \mbox{Computer} Science: cache invalidation and naming things. - Phil Karlton}
\end{quote}
This famous quote refers to the innate tendency of computer scientists to choose poor names for a program's variables despite the existence of good coding practices \citep{mcconnell_code_2004}. Practically speaking, this is not a problem as long as the program does its job, but can we still argue the same in the era of Large Language Models (LLMs)?

LLMs \citep{brown_language_2020} have significantly changed how we process text by providing a straightforward interface, i.e., natural language, to define the problem to be solved \citep{devlin_bert_2019}. They provide software engineers with useful coding tips and can be used as part of larger and more complex software systems \citep{dakhel_github_2023}. It is common to set up an LLM using a set of instructions called ``prompt'', and it soon became clear to both researchers and developers that the prompt itself can greatly influence an LLM's performance \citep{zhao_calibrate_2021,sclar_quantifying_2024,yang_large_2024}. The process of writing a good prompt for the current task is called prompt engineering, and a great deal of different techniques have been proposed in this direction \citep{nori_can_2023,sahoo_systematic_2024}, ranging from a simple description of the problem to few-shot examples.

\begin{figure*}[t]
    \centering
    \resizebox{1\textwidth}{!}{\input{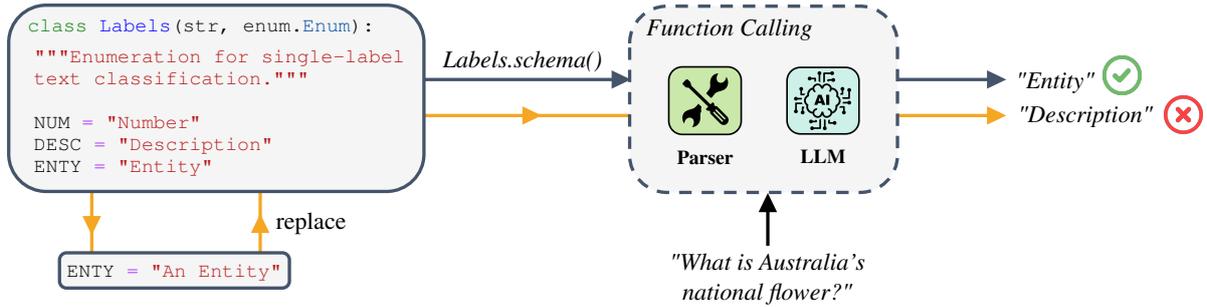}}
    \caption{Example of GPT3.5 behavior when classifying a question in terms of what it is referring to. A slight change in the definition of the class ``ENTY'' causes a minor prompt variation that disrupts the LLM's prediction. This happens under the hood, making it very hard for a developer to debug the program. Note that the same might happen, for instance, if the ordering or naming of variables is changed (hence the quote of Section \ref{sec:introduction}).}
    \label{fig:motivation}
\end{figure*}

From a practical standpoint, integrating LLMs in a software infrastructure introduces additional complexities, from choosing good prompt engineering strategies to parsing and controlling the output format of responses. In an effort to simplify the LLM into a straightforward function call, new software libraries like Instructor \citep{liu_2024_instructor} have emerged, reducing even further the entry barrier of LLMs for the ordinary programmer. Instructor requires to define target labels as static fields of a Python class, which is then automatically converted into a standardized JSON schema. The schema and the input sentence are then used to produce an LLM response via a mechanism known as Function Calling \citep{kang_preliminary_2023}. Figure \ref{fig:motivation} provides an example of text classification using Instructor. Here, the assumption of the programmer \textit{should} be that small changes to the labels' descriptions, as well as the labels' ordering in the code, should not affect the final result. Unfortunately, reality is different: a change in the label definition, such as adding an article as shown in the figure, or in variables' naming can lead to minor prompt variations with drastic changes in the final prediction. As a result, developers remain unaware of the malfunction's cause and might later abandon the tool due to frustration. 

This simple yet troubling example might be generalized to other problems, for instance, code generation \citep{liu_your_2023}; at the end of the day, LLM engineers would like to know whether it makes sense to spend their time modifying prompts to obtain a better-behaving LLM; perhaps the LLM does not change its predictions much no matter how the prompt is re-written. Similarly, an LLM whose predictions greatly vary depending on how the prompt is written might generally be regarded as unreliable in a production environment. Therefore, the question we want to address in this paper is the following: \textit{"how can we quantify the sensitivity of an LLM to variations of the prompt?"}.
Existing works have answered this question by considering accuracy as the sole metric of interest \citep{mccoy_embers_2023}, but this has a limited impact on the everyday life of developers and requires enough ground truth labels for the estimate to be reliable. As a matter of fact, with the recent progress in LLM agents \citep{gioacchini_agentquest_2024} and chain of thoughts \citep{wei_chain_2022} techniques, the existence of multiple intermediate steps and/or user inputs, each handled somehow by an LLM, implies an exponential amount of potential failure paths. LLM engineers need a computationally feasible way to analyze each step individually, possibly irrespective of the final task they need to solve, to reduce the chances that something goes wrong along the way.

\begin{figure}[t]
    \centering
    \includegraphics[width=0.9\linewidth]{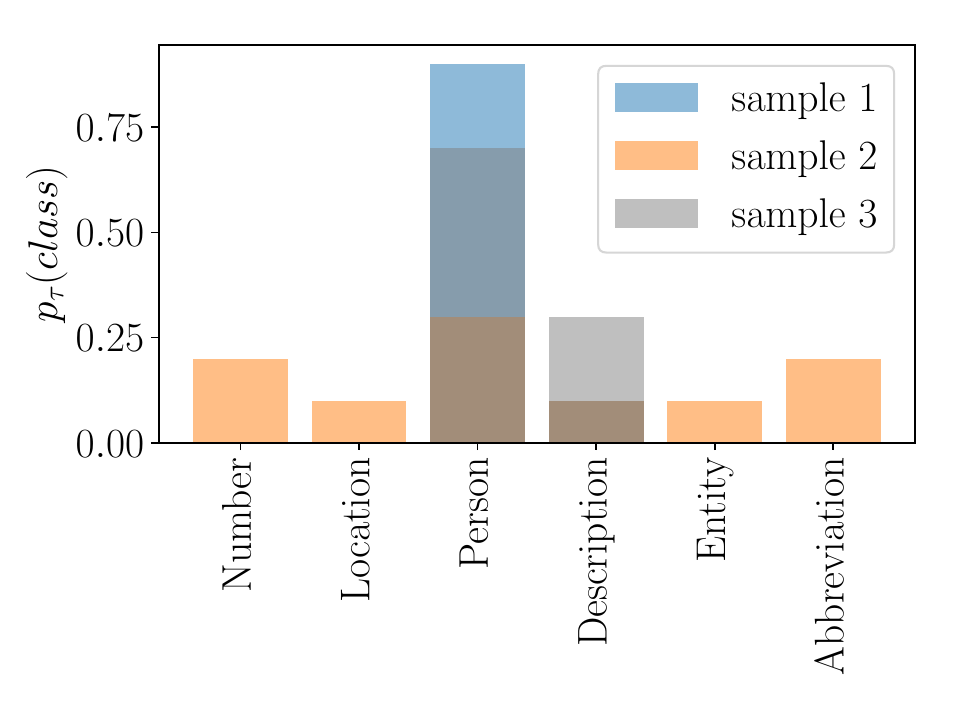}
    \caption{Predicted class distributions over prompt rephrasings $p_\tau$ across three samples of the same class Person (TREC dataset, Section \ref{sec:experiments}). Merely syntactic prompt rephrasings can produce very diverse distributions. For instance, sample 2 is characterized by \textit{high sensitivity} and (compared to others) \textit{low consistency}.}
    \label{fig:example-swap}
\end{figure}

We set out to address these problems by proposing two diagnostic metrics for black-box classifiers that are complementary to accuracy: the \textit{sensitivity} 
to the input, which does not depend on the ground truth labels, and the \mbox{\textit{consistency}} of predictions across examples of the same class. Intuitively, it is desirable to have LLMs that are robust to semantically equivalent variations of the initial prompt, and their predicted labels' distributions should not vary much across samples of the same class. 
Striving to improve these two metrics towards \textbf{low sensitivity} and \textbf{high consistency} might significantly reduce the unpredictability of LLMs' behavior in complex software systems running in production. 
Consider the illustrative example of Figure \ref{fig:example-swap}, where three samples of the same class Person are classified (in probability) very differently when testing different variations of the same prompt. Samples 1 and 3 have lower sensitivity than sample 2, as their distributions are quite stable on the correct class. In addition, sample 2 has low consistency compared to the other samples, which combined with sensitivity indicates ``hardness'' of classification. The behavior of sample 2 is highly undesirable, and the first step to address it is to formalize metrics that reflect these intuitions.

We empirically test these metrics, finding that they indeed convey different information about the LLMs' behavior. We use this information to qualitatively analyze some of the datasets, mimicking what an LLM engineer would do when debugging a real system. We also show how these metrics help to improve prompts by identifying different sets of problematic samples.
\section{Related Work}
\label{sec:related-work}
This section positions our work in the context of three different but related research directions: influence of spurious features, uncertainty quantification, and prompt optimization. All these works fall under the broader umbrella of prompt engineering.

\paragraph{Spurious Features.}
It is well-known that variations of the prompt affect stability of the LLM's accuracy. In a broad study across diverse tasks, \citet{mccoy_embers_2023} showed that LLM performances depend on the likelihood of the input prompt and of the correct output answer. Such result is consistent with LLMs' autoregressive nature, that is, models trained to maximize a likelihood objective. In light of these considerations, it is not surprising that changing the ordering of the examples in a few-shot prompting strategy can lead to almost random accuracy on sentiment analysis tasks \citep{zhao_calibrate_2021}, or that the nature of prompts affects LLMs' benchmarks \citep{ailem_examining_2024}. Similar considerations motivated frameworks like FormatSpread, which predicts the expected performance under prompt's variations without accessing the LLM's weights \citep{sclar_quantifying_2024}. Also, spurious features in the prompt have severe repercussions on security vulnerabilities' detection, where LLMs appear inconsistent and unfaithful \citep{ullah_llms_2024}, while \citet{yang_just_2024} study the effect of prompt rephrasings and LLMs' temperature on classification and uncertainty metrics. Finally, we mention the comprehensive benchmark of \cite{liang_holistic_2023}, where metrics such as invariance to semantic-preserving transformations are computed w.r.t.\ accuracy changes. Note that the solution to the spurious feature problem might also reside in a more structured approach: Retrieval Augmented Generation \citep{lewis_retrieval_2020} or Knowledge Graphs-enhanced LLMs \citep{luo_reasoning_2024} reduce hallucinations and prompt dependencies. 

\paragraph{Uncertainty Quantification.} Complementary to the discussion in this paper is the estimation of the LLM's uncertainty. Typically, uncertainty is defined over the different answers of the LLM given the \textit{same} prompt \citep{press_entropy_2024}; being generative models of text, LLMs might produce different predictions due to stochasticity in the output response. Several works have already investigated uncertainty: \citet{kadavath_language_2022} ask the LLM to provide a score of its confidence, whereas \citet{chen_more_2024} argued that a majority voting mechanism reveals a non-monotonic relationship between the number of LLM calls and system performance. Motivated by the fact that higher uncertainty should imply lower performances, \citet{huang_uncertainty_2024} define a rank calibration error to quantify deviations from the ideal relationship between the two quantities. In a human study with about 400 participants, \citet{kim_notsure_2024} provide evidence that an LLM expressing uncertainty in natural language reduces the users' trust in the system. Very recently, \citet{yadkori_believe_2024} proposed an information-theoretic metric to distinguish between epistemic and aleatoric uncertainty. They enable the identification of unreliable model outputs and hallucinations without altering the training process.

It does not fall within the scope of this paper to analyze output uncertainty. Because LLMs can also behave pseudo-deterministically \citep{yang_just_2024}, this limits the impact of some probability-based uncertainty quantification metrics. For a comprehensive study of uncertainty quantification techniques on LLMs, such as verbalization and generation likelihood, we refer the reader to \citet{huang_look_2023} and \citet{geng_survey_2024}.

\paragraph{Prompt Optimization.}
To improve LLMs performance, one can carefully engineer their prompt. One cause for LLMs' bad performances is the bias towards over-represented classes; for this reason, \citet{zhao_calibrate_2021} propose a calibration technique that makes the predictions more uniform across classes. The approach requires access to the LLM's inner workings, which might be impractical. Automated Prompt Engineer \citep{zhou_large_2023} optimizes prompts with the help of an LLM, reducing its susceptibility to adversarial attacks and jailbreaks that overcome the safeguards for ethical use of these systems \citep{zhou_robust_2024}. At the same time, there is still much work to do: LLM-based automatic prompt optimizers struggle to identify the true causes of errors, and we should rather focus on an automated \textit{behavior} optimization paradigm \citep{ma_large_2024}. Prompt optimization is already a step ahead compared to the quantification of our metrics, and we hope that this work will inspire new prompting strategies that do not focus on the sole evaluation of task accuracy.
\section{Methodology}
\label{sec:method}
We formally introduce the metrics of interest, sensitivity and consistency, in a bottom-up fashion.

Let us consider a classification task $\tau$ with $C$ possible classes and a data set of $N$ i.i.d.\ samples $\mathcal{D}={(\bm{x}_i, y_i)}_{i \in \{1,\dots, N\}}$, with $\bm{x}$ being the input text to classify and \mbox{$y \in \{1,\dots,C\}$} the ground truth label. Moreover, we refer to the subset of samples of class $y$ as $\mathcal{D}_y$. We make the typical conditional independence assumption about the data, that is, $p(y,\bm{x})=p(y|\bm{x})p(\bm{x})$. Without loss of generality, we introduce the distribution $p_{\rho_0}(\rho)$ defined over all prompts that are semantically equivalent to a reference prompt $\rho_0$, which is typically prepared by the user based on task $\tau$; when clear from the context, we will omit the subscript from the notation. Because this distribution depends on the task but is independent of the input distribution, we shall write $p(y,\bm{x},\rho) = p(y|\bm{x},\rho)p(\bm{x})p(\rho)$. Please note that the distribution $p(\rho)$ is not uniform. In practice, LLMs implement the distribution $p(y|\bm{x},\rho)$, hence we can define the average probability that an LLM predicts the class $y$ under different variants of the same prompt as
\begin{align}
    p_\tau(y|\bm{x}) = \mathbb{E}_{\rho \sim p(\cdot)}[p(y|\bm{x},\rho)],
\label{eq:llm-categorical-over-prompts}
\end{align}
which is different from the class distribution predicted by classifiers for a fixed $\bm{x}$. Similarly to $p(\bm{x})$, $p(\rho)$ is unknown; therefore, we can approximate Equation \ref{eq:llm-categorical-over-prompts} using a sampler $\mathcal{S}$ \citep{barber_bayesian_2012} over $Q$ semantically equivalent prompts $\rho_1,\dots,\rho_Q$\footnote{$\rho_1=\rho_0$ in our experiments, that is, the original question is one of the prompts.}:
\begin{align}
    p_\tau(y|\bm{x}) \approx \frac{1}{Q}\sum_{i=1}^Q p(y|\bm{x},\rho_i).
\label{eq:monte-carlo-approximation}
\end{align}
In this work, the sampler $\mathcal{S}$ is an LLM that is tasked to rephrase prompts (Appendix \ref{sec:prompt-rephrasing} shows examples), but it could also be an algorithm that generates text modifications as in \citet{liang_holistic_2023}.
We then define the sensitivity of an LLM to the prompt to reflect how much the LLM prediction varies under the rephrasings of the original prompt.
\begin{definition}[Sensitivity] 
Given a distribution $p_\tau(y|\bm{x})$ defined as in Equation \ref{eq:llm-categorical-over-prompts}, the sensitivity with respect to $\bm{x}$ is the normalized entropy
\begin{align}
    S_\tau(\bm{x}) = -\mathbb{E}_{y \sim p_\tau(\cdot|\bm{x})}[\ln p_\tau(y|\bm{x})]/\ln(C),    
\end{align}
whereas the expected sensitivity is
\begin{align}
    S_\tau = \mathbb{E}_{\bm{x}}[S_\tau(\bm{x})] \approx \frac{1}{N} \sum_{i=1}^N S_\tau(\bm{x}_i).  
    \label{eq:sensitivity}
\end{align}
\end{definition}
It is important to note that the sensitivity does not require access to ground truth labels, which are often hard to acquire, and it does not necessarily correlate with the task's performance. The sensitivity should be used, for instance, as a guide to compare the ``robustness'' of different LLMs to variations of the prompt. A highly \textit{sensitive} LLM may require significant prompt optimization efforts, whereas a less sensitive LLM tells us there might be no further room for improvement. The \textit{scope} of the prompt variation is also important: one can measure sensitivity w.r.t. minor variations of the original prompt, as well as completely different prompting strategies as long as they convey a semantically equivalent instruction to the LLM. The interpretation we attribute to sensitivity ultimately depends on the use case, but it is universal.  

The second metric is called consistency. It measures how much the distribution of Equation \ref{eq:llm-categorical-over-prompts} differs for two samples $\bm{x},\bm{x}'$ of the same class $y$ using the Total Variation Distance (TVD):
\begin{align}
    \text{TVD}(p,q) = \frac{1}{2}\sum_{c=1}^C |p(c) - q(c)|,
\label{eq:total-variation-distance}
\end{align}
whose values range between $0$ and $1$.
\begin{definition}[Consistency] 
Given a categorical distribution $p_\tau(y|\bm{x})$ defined as in Equation \ref{eq:llm-categorical-over-prompts} and two samples $\bm{x},\bm{x}' \in \mathcal{D}_y$, the pair-wise consistency of a classifier is measured as
\begin{align}
    C_y(\bm{x},\bm{x}') = 1-\textnormal{TVD}(p_\tau(\cdot|\bm{x}), p_\tau(\cdot|\bm{x}')],    
\end{align}
whereas the expected consistency is
\begin{align}
    C_y = \mathbb{E}[C_y(\bm{x},\bm{x}')] \approx \sum_{\bm{x},\bm{x}' \in \mathcal{D}_y} \frac{C_y(\bm{x},\bm{x}')}{|\mathcal{D}_y|^2}.    
    \label{eq:consistency}
\end{align}
\end{definition}
Intuitively, a \textit{consistent} LLM produces similar distributions $p_\tau(\cdot|\bm{x})$ regardless of the sample $\bm{x}$ of class $y$. When sensitivity is not 0, being consistent suggests that prompt rephrasings cause similar mistakes across all samples of class $y$, hence a careful tuning of the prompt is required. Instead, an inconsistent LLM behaves unpredictably among samples of the same class, where the same prompt rephrasings cause different mistakes; this might indicate that the problem is not the prompt, rather the classifier itself. Note that, when sensitivity is 0, (in)consistency is uniquely determined by the inherent difficulty of the classification task, as it counts pair-wise class mismatches in the LLM predictions. In this case, consistency will correlate more with classification metrics such as accuracy, thus we expect it to be most useful when sensitivity is high enough, e.g., $0.1$.

We argue that, in order to avoid bad surprises in production environments where new LLMs have to be tested and replaced quickly, it might be desirable to select LLMs with \textbf{low sensitivity} and \textbf{high consistency}, which means $S_\tau \rightarrow 0$ and $C_y \rightarrow 1, \forall y \in \{1,\dots,C\}$. 

Finally, consider how probability-based output uncertainty is defined by \citet{huang_uncertainty_2024} for \textit{fixed} input $\bm{x}$ and prompt $\rho$:
\begin{align}
    U(\bm{x}, \rho) = -\mathbb{E}_{y \sim p(\cdot|\bm{x},\rho)}[\ln p(\cdot|\bm{x},\rho)].
\end{align}
Output uncertainty is orthogonal to sensitivity, as it does not focus on variations of the prompt $\rho$. 
\section{Experiments}
\label{sec:experiments}
The goal of the experiments is to assess how sensitivity and consistency can help in the context of English-based text classification datasets, analyzing the impact of prompt variations and prompting strategies from a different angle compared to previous works, but most importantly showing developers how to use these metrics\footnote{The code is available at: \url{https://github.com/nec-research/sensitivity-consistency-LLM}.}. Therefore, it is outside the scope of this work to perform an extensive benchmark of these metrics across many datasets and LLMs, which would not be helpful to our purpose. We ran the LLMs on a 4 Tesla V100 server with 32 GBs of memory and 252 GBs of RAM.

We consider \textit{Llama-3-70B-Instruct} \citep{touvron_llama_2023} and \textit{Mixtral-8x7B-Instruct-v0.1} \citep{jiang_mixtral_2024} as the two open-source LLMs available on our servers, as well as \textit{GPT-3.5-turbo-0125} and \textit{GPT-4o-2024-08-06} \citep{brown_language_2020} as closed-source models. The temperature is set to zero and the seed is fixed to 42 to obtain quasi-deterministic behavior. Following \citet{zhao_calibrate_2021}, five multiclass classification datasets are used for the comparison: \textbf{TREC}, a 6-class (and 50 subclasses) question-answering task \citep{voorhees_building_2000} with 500 test samples, CommittmentBank (\textbf{CB}) as a 3-way classification problem \citep{de_commitmentbank_2019} with 250 test samples, a binary textual entailment problem (\textbf{RTE}, \citet{dagan_pascal_2005}) with 2490 test samples, and the 14-class ontology extraction dataset \textbf{DBPedia} \citep{zhang_character_2015} with 2000 balanced test samples. In addition, we also consider the 7-class Web of Science 46985 (\textbf{WoS}) dataset with 2000 balanced test samples \citep{kowsari_web_2018}. In the event the LLM cannot produce a valid class, we add an extra class label \textit{N/A}.

We analyze three different prompting strategies: a \textit{simple} strategy, where the prompt consists of the \textbf{task description} and the list of classes; a \mbox{\textit{detail}} strategy, where we provide a detailed description of each class; and a \textit{1-shot} strategy in which, compared to \textit{simple}, we also provide one example for each class taken from samples that do not belong to the test set.
To build different rephrasings of the task description \textit{only}, we use the aforementioned LLMs. In particular, the prompt asks to rephrase the task description by changing the length or adding unnecessary words as long as the meaning remains the same. For the purpose of our work, which is only to show how sensitivity and consistency can be used, we fix $Q=30$ after observing that increasing this value did not vary results significantly and was enough for our demonstrative purposes. Also, the ``right'' value of $Q$ ultimately depends on the user's compromise between computational resources and stability of resulting statistics. In the quantitative analyses, we display the average sensitivity and average consistency, as well as the micro F1-score across all $\rho$. Sensitivity and micro F1-score are averaged across samples, whereas pair-wise consistencies are averaged together. Since the distributions of sensitivity and consistency values are far from being Gaussian, looking at certain statistics such as the standard deviation can convey a misleading message of instability. Therefore, we will perform qualitative analyses in later sections that show large deviations from the mean values, but in the interest of completeness we report standard deviations in Appendix \ref{sec:std-results}. 
\section{Results}
\label{sec:results}
In this section, we want to answer two questions: \\ \textbf{i)} are sensitivity and consistency complementary to accuracy metrics?; \textbf{ii)} how can we use them to fix prompts, LLMs, and choose the most suitable LLMs for a specific use case?

\begin{table*}[t]
\small
\begin{tabular}{lcccccc}
\toprule
                     & \multicolumn{3}{c}{Llama3}                                                                                                                                                                                                       & \multicolumn{3}{c}{Mixtral}                                                                                                                                                                                                      \\ \midrule
\multicolumn{1}{c}{} & \begin{tabular}[c]{@{}c@{}}Simple\\ $S_\tau$/$C_y$/F1\end{tabular} & \begin{tabular}[c]{@{}c@{}}Detail\\ $S_\tau$/$C_y$/F1\end{tabular} & \multicolumn{1}{c}{\begin{tabular}[c]{@{}c@{}}1-shot\\ $S_\tau$/$C_y$/F1\end{tabular}} & \begin{tabular}[c]{@{}c@{}}Simple\\ $S_\tau$/$C_y$/F1\end{tabular} & \begin{tabular}[c]{@{}c@{}}Detail\\ $S_\tau$/$C_y$/F1\end{tabular} & \multicolumn{1}{c}{\begin{tabular}[c]{@{}c@{}}1-shot\\ $S_\tau$/$C_y$/F1\end{tabular}} \\ \midrule
TREC    & .127/.693/{\color{OrangeRed}\textbf{.848}} & .141/.694/.824 & {\color{OrangeRed}\textbf{.095}}/.686/.846 & .223/.657/.736 & .182/{\color{OrangeRed}\textbf{.709}}/.728 & .166/.640/.744 \\
CB      & .018/.956/.920 & .016/{\color{OrangeRed}\textbf{.963}}/.924 & .014/.962/{\color{OrangeRed}\textbf{.936}} & .203/.625/.672 & {\color{OrangeRed}\textbf{.013}}/.695/.788 & .017/.600/.652 \\
RTE     & .157/.701/.569 & .213/.727/.509 & .048/{\color{OrangeRed}\textbf{.755}}/.818 & .195/.669/.814 & .176/.676/.756 & {\color{OrangeRed}\textbf{.046}}/.754/{\color{OrangeRed}\textbf{.847}} \\
DBPedia & {\color{OrangeRed}\textbf{.018}}/{\color{OrangeRed}\textbf{.932}}/{\color{OrangeRed}\textbf{.948}} & .058/.901/.740 & .030/.924/.827 & .053/.802/.804 & .054/.808/.862 & .069/.773/.819 \\
WoS     & .043/.479/{\color{OrangeRed}\textbf{.627}} & {\color{OrangeRed}\textbf{.034}}/{\color{OrangeRed}\textbf{.480}}/.623 & .030/.468/.580 & .112/.483/.606 & .124/.469/.599 & .142/.472/.566 \\ \bottomrule
\end{tabular}
\begin{tabular}{lcccccc}
\toprule
                     & \multicolumn{3}{c}{GPT-3.5}                                                                                                                                                                                                       & \multicolumn{3}{c}{GPT-4o}                                                                                                                                                                                                      \\ \midrule
\multicolumn{1}{c}{} & \begin{tabular}[c]{@{}c@{}}Simple\\ $S_\tau$/$C_y$/F1\end{tabular} & \begin{tabular}[c]{@{}c@{}}Detail\\ $S_\tau$/$C_y$/F1\end{tabular} & \multicolumn{1}{c}{\begin{tabular}[c]{@{}c@{}}1-shot\\ $S_\tau$/$C_y$/F1\end{tabular}} & \begin{tabular}[c]{@{}c@{}}Simple\\ $S_\tau$/$C_y$/F1\end{tabular} & \begin{tabular}[c]{@{}c@{}}Detail\\ $S_\tau$/$C_y$/F1\end{tabular} & \multicolumn{1}{c}{\begin{tabular}[c]{@{}c@{}}1-shot\\ $S_\tau$/$C_y$/F1\end{tabular}} \\ \midrule
TREC    & .128/.573/.710 & .148/.549/.632 & .100/.649/.788 & .042/.797/.848 & .068/.708/.788 & {\color{OrangeRed}\textbf{.037}}/{\color{OrangeRed}\textbf{.816}}/{\color{OrangeRed}\textbf{.862}} \\
CB      & .274/.730/.720 & .330/.732/.636 & .288/.737/.716 & .087/.814/.876 & .081/.826/.884 & {\color{OrangeRed}\textbf{.058}}/{\color{OrangeRed}\textbf{.880}}/{\color{OrangeRed}\textbf{.944}} \\
RTE     & .292/.719/.780 & .346/.719/.781 & .218/.703/.743 & .233/.757/.878 & .087/.823/.889 & {\color{OrangeRed}\textbf{.074}}/{\color{OrangeRed}\textbf{.835}}/{\color{OrangeRed}\textbf{.916}} \\
DBPedia & .026/.916/.938 & .028/.895/.899 & .026/.914/{\color{OrangeRed}\textbf{.954}} & .023/{\color{OrangeRed}\textbf{.939}}/.920 & .022/.921/.914 & {\color{OrangeRed}\textbf{.017}}/.922/.946 \\
WoS     & .095/.482/.635 & .093/.480/.622 & .120/.486/.665 & .048/.492/.665 & {\color{OrangeRed}\textbf{.045}}/.493/.666 & .046/{\color{OrangeRed}\textbf{.494}}/{\color{OrangeRed}\textbf{.668}}\\ \bottomrule
\end{tabular}
\caption{Sensitivity $S_\tau$ (lower is better), average consistency $C_y$ across all sample pairs (higher is better), and micro-F1 score across all prompt rephrasings $\rho$ are shown for different datasets, models, and prompting strategies. Best values across open and closed-source models are shown in bold. These results only support the demonstration of practical utility of the proposed metrics and do not serve as an extensive benchmark across datasets and models.}
\label{tab:quantitative-results}
\end{table*}

\paragraph{Quantitative Results} To answer question \textbf{i)}, Table \ref{tab:quantitative-results} reports, for each model, dataset, and prompting strategy tried, the values of sensitivity, consistency, and micro F1 score. The first observation is that there seems to be no consistent agreement between the proposed metrics across open-source LLMs and prompting strategies when sensitivity is sufficiently high, whereas GPT-4o with the 1-shot strategy shows excellent performances in all tasks. When sensitivity is close to zero, $p_\tau$ collapses and the influence of $Q$ vanishes; this is straightforwardly associated to low/high consistency if the classifier is bad/good, respectively. To further support these arguments, the Pearson correlation between sensitivity and consistency (without said degenerate cases where sensitivity is less than 0.05) has a value of \textit{-0.07}.
These results agree with our intuitions about the utility of these metrics: a random predictor $p_\tau(y|\bm{x})$ would achieve $S_\tau\approx1$ but $C_y\approx1$, whereas another that always predicts a specific class has $S_\tau\approx0$ and $C_y\approx1$. All combinations are possible, providing different views about LLMs' behavior.

As a result, developers should pay attention when switching LLMs in their applications: a prompt that worked well with an LLM might cause instability (see the sensitivity gaps on CB) and significantly worse performance on another. For instance, on CB it might be preferable to choose the Detail strategy with Mixtral, since sensitivity is extremely low and the other metrics are highest compared to the alternative strategies. To further validate our results, Appendix \ref{sec:comaprison-noise-random-predictors}, provides evidence of the significant deviation of our results compared to perturbed and random predictors.

\vspace{0.2em}
\take{Sensitivity and consistency convey distinct information, especially when sensitivity is high enough (>0.05).}

\paragraph{Sensitivity Analysis}

\begin{figure*}[t]
    \centering
        \begin{subfigure}{0.329\textwidth}
        \centering
        \includegraphics[width=\textwidth]{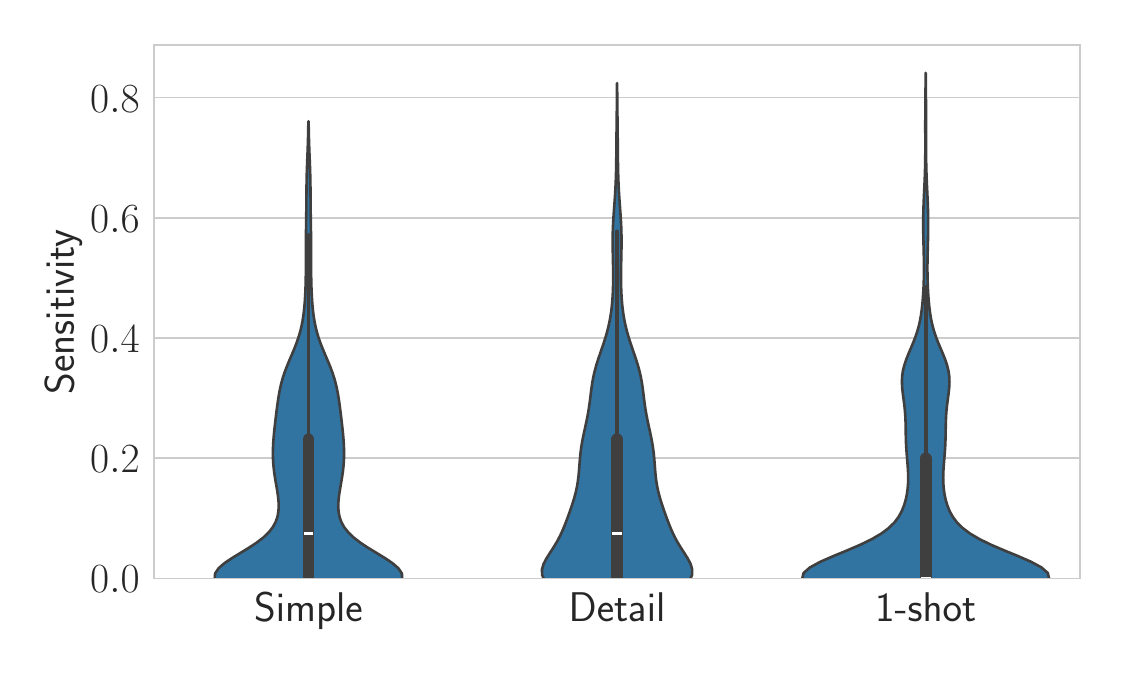}
        \caption{TREC}
    \end{subfigure}
    \begin{subfigure}{0.329\textwidth}
        \centering
        \includegraphics[width=\textwidth]{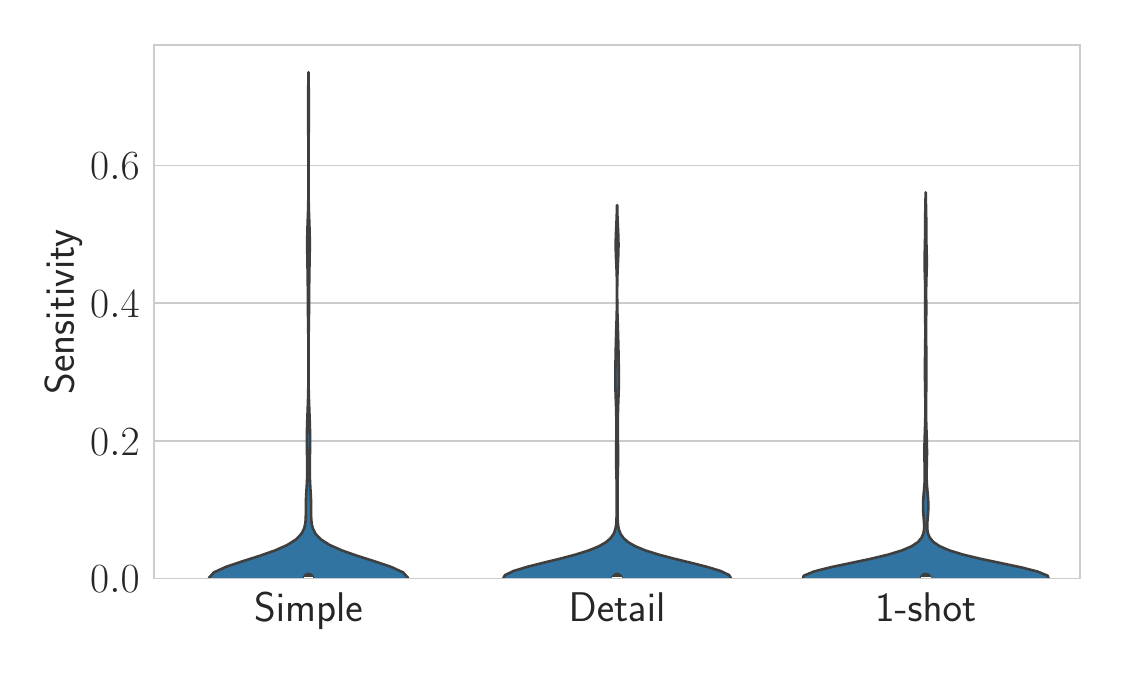}
        \caption{CB}
    \end{subfigure}
    \begin{subfigure}{0.329\textwidth}
        \centering
        \includegraphics[width=\textwidth]{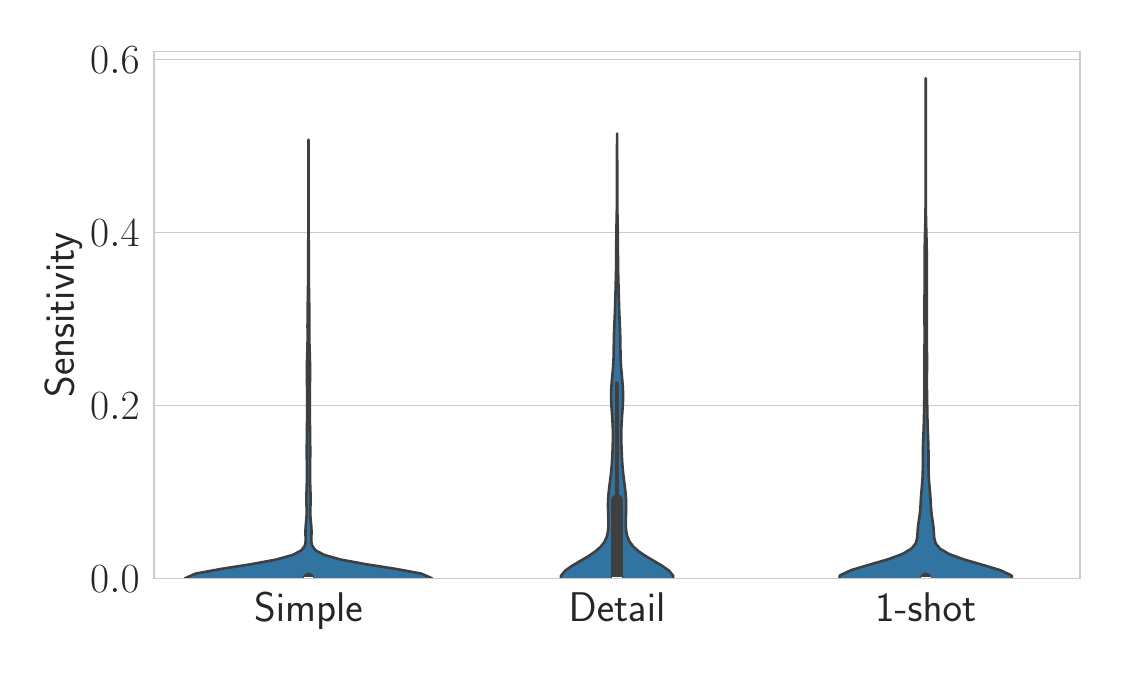}
        \caption{DBPedia}
    \end{subfigure}
    \begin{subfigure}{0.329\textwidth}
        \centering
        \includegraphics[width=\textwidth]{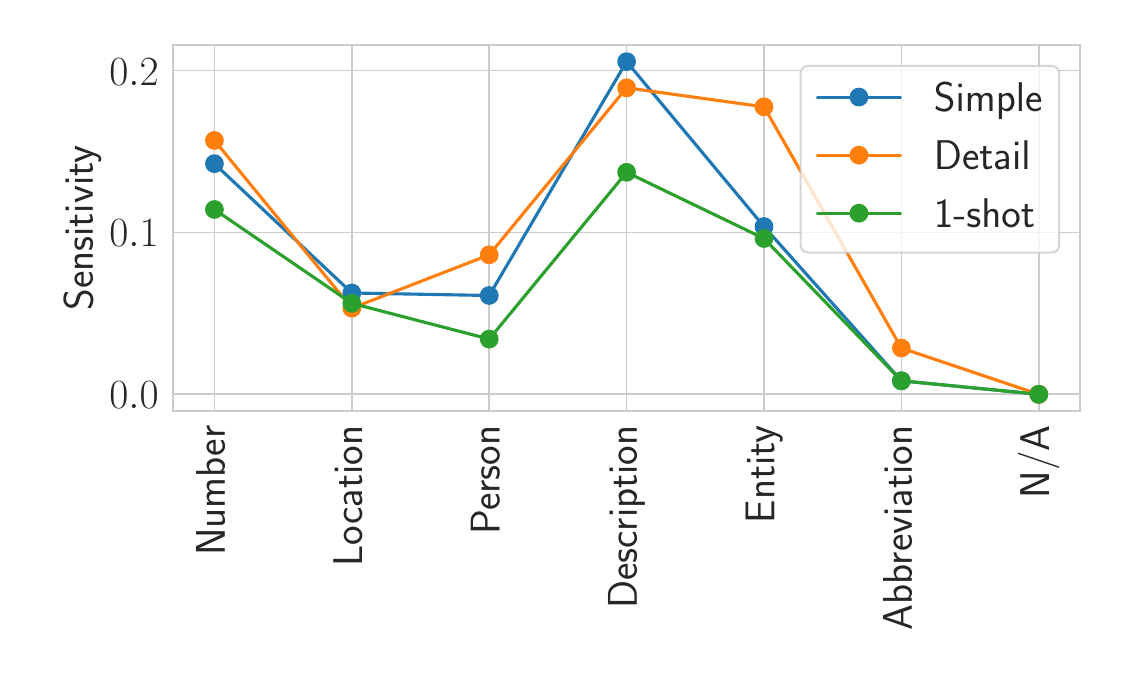}
    \end{subfigure}
    \begin{subfigure}{0.329\textwidth}
        \centering
        \includegraphics[width=\textwidth]{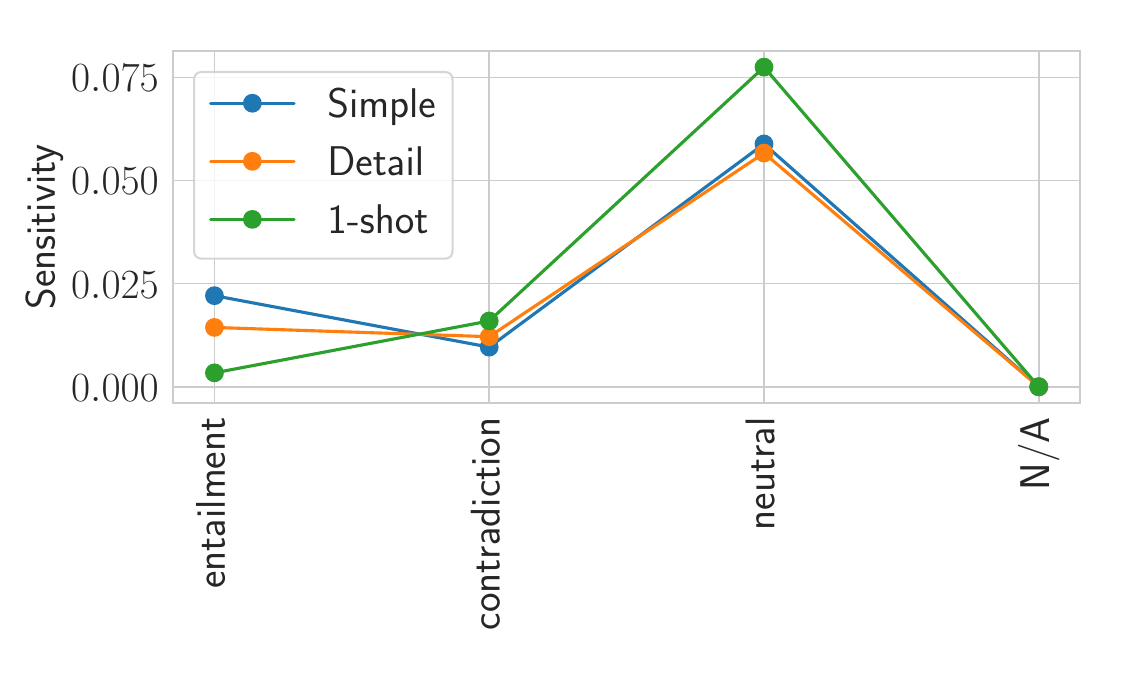}
    \end{subfigure}
    \begin{subfigure}{0.329\textwidth}
        \centering
        \includegraphics[width=\textwidth]{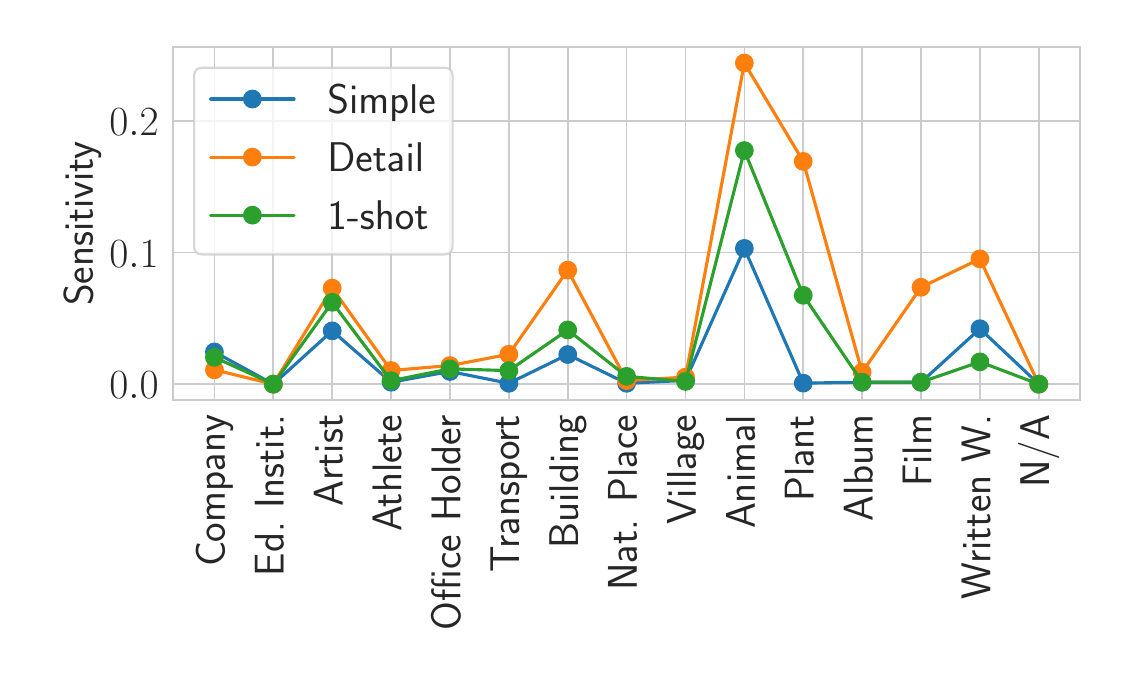}
    \end{subfigure}
    \caption{\textit{Top:} We show the sensitivity for each sample of the dataset according to different prompting strategies. \textit{Bottom:} we plot the sensitivity $S_\tau$ for each class and prompting strategy (Llama3). We remind that the prompting strategy itself might be considered another semantically equivalent rephrasing of the initial prompt $\rho_0$ (Section \ref{sec:method}).}
    \label{fig:sensitivity-analysis}
\end{figure*}

We now answer question \textbf{ii)}, by inspecting the sensitivity values against each class and prompting strategy. In Figure \ref{fig:sensitivity-analysis} (top), we visualize the distributions of sensitivity across samples and dataset, divided by prompting strategy. These distributions help us understand the behavior of the LLM compared to simply checking mean values. For instance, Llama3 has low average TREC sensitivity with the 1-shot strategy, but there are a non-negligible number of samples for which sensitivity is still very high. \textit{If a developer does not have access to ground truth labels at all}, these samples can be manually inspected or given to labelers to identify situations that need special care. As a result, the labeling cost is reduced and the developer knows how to improve the prompt. In Appendix \ref{sec:sensitivity-improve-prompts}, we demonstrate how to adjust prompts according to most sensitive samples, which reduces sensitivity and improves classification accuracy.

When some ground truth information is available, Figure \ref{fig:sensitivity-analysis} (bottom) can give additional insights into the problematic classes of the task. Consistently with the example of Section \ref{sec:introduction}, we see that on TREC the classes Description, Entity, and Number are the ones with the highest sensitivity. This result reveals that the LLM is unsure about samples that we -- as human evaluators -- also found ambiguous to classify when looking at the data, especially as regards Description and Entity classes. A similar result holds for CB, where the neutral statements are the hardest to classify; here Llama3 is more sensitive to rephrasings of the task description when it comes to neutral statements. This suggests that it might be worth increasing the number of few-shot examples for that class, for instance, or providing a better definition of a neutral statement in the prompt. On DBPedia, instead, it seems that Llama3 has sensitivity issues with the samples of class Artists Building, Animal, and Written Work.

\vspace{0.2em}
\take{Sensitivity can be used with or without ground truth labels to find ``problematic'' samples, revealing LLMs' weak spots.}

\paragraph{Consistency Analysis}

\begin{figure*}[t]
    \centering
    \begin{subfigure}{0.329\textwidth}
        \centering
        \includegraphics[width=\textwidth]{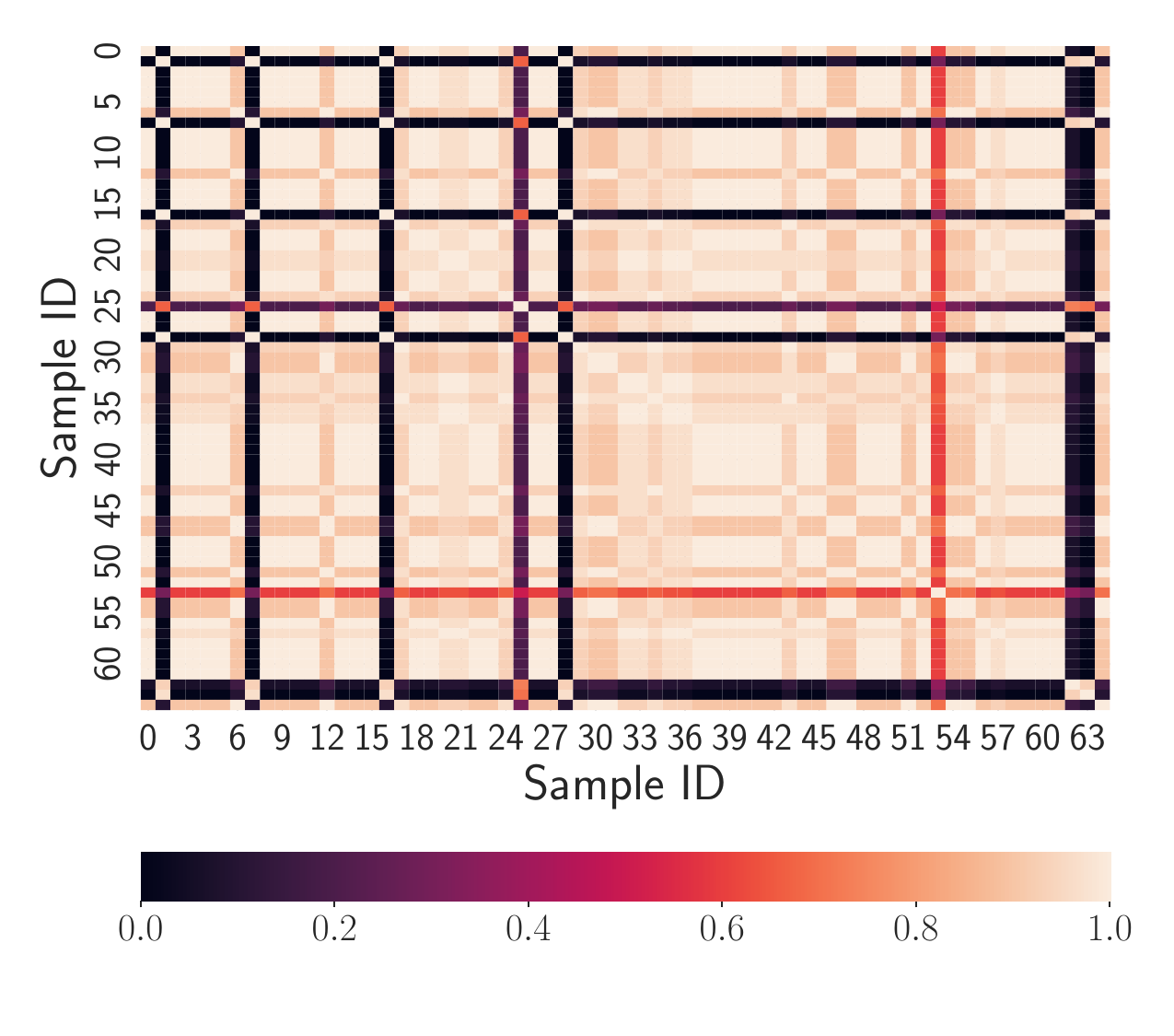}
    \end{subfigure}
    \begin{subfigure}{0.329\textwidth}
        \centering
        \includegraphics[width=\textwidth]{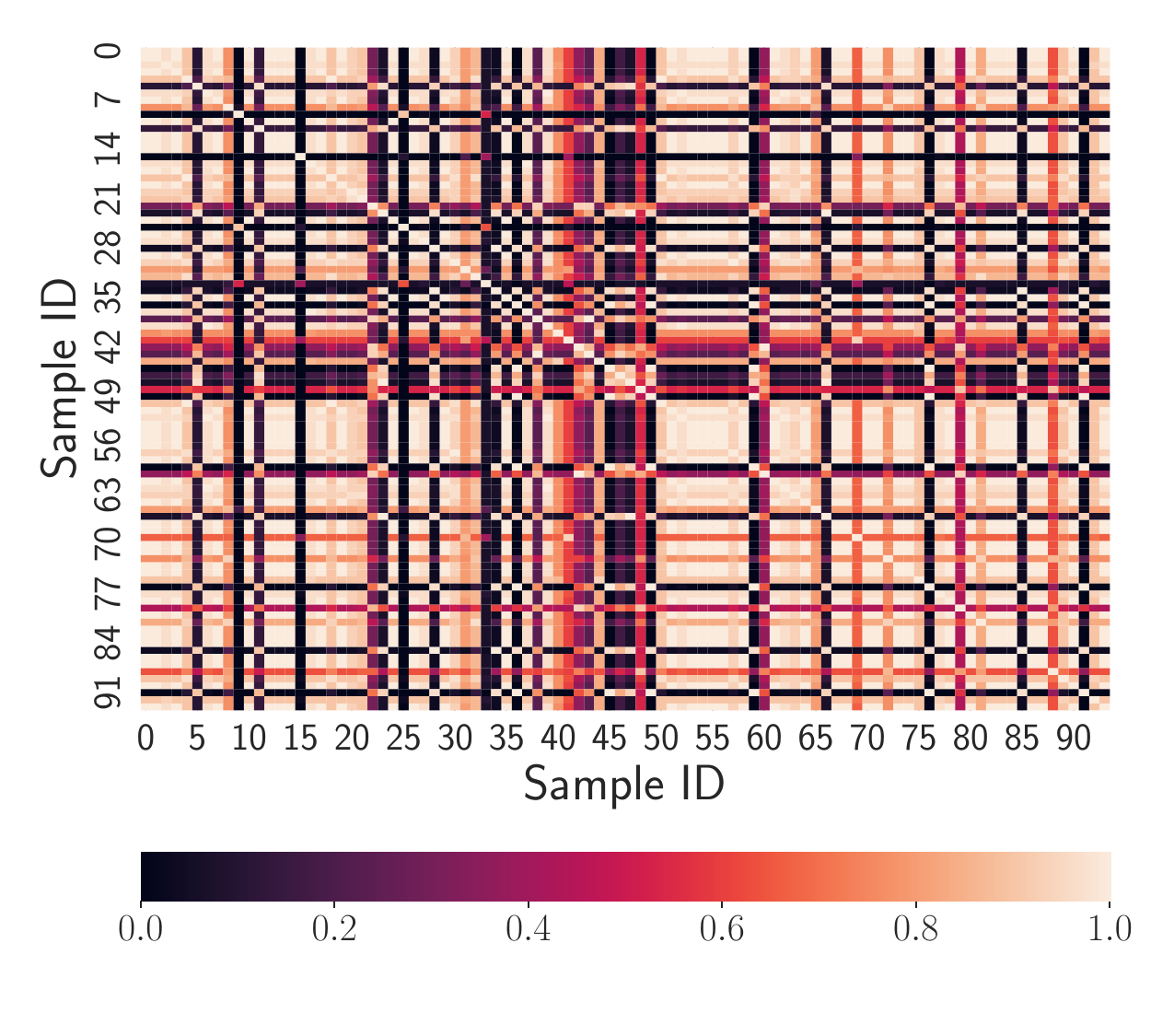}
    \end{subfigure}
    \begin{subfigure}{0.329\textwidth}
        \centering
        \includegraphics[width=\textwidth]{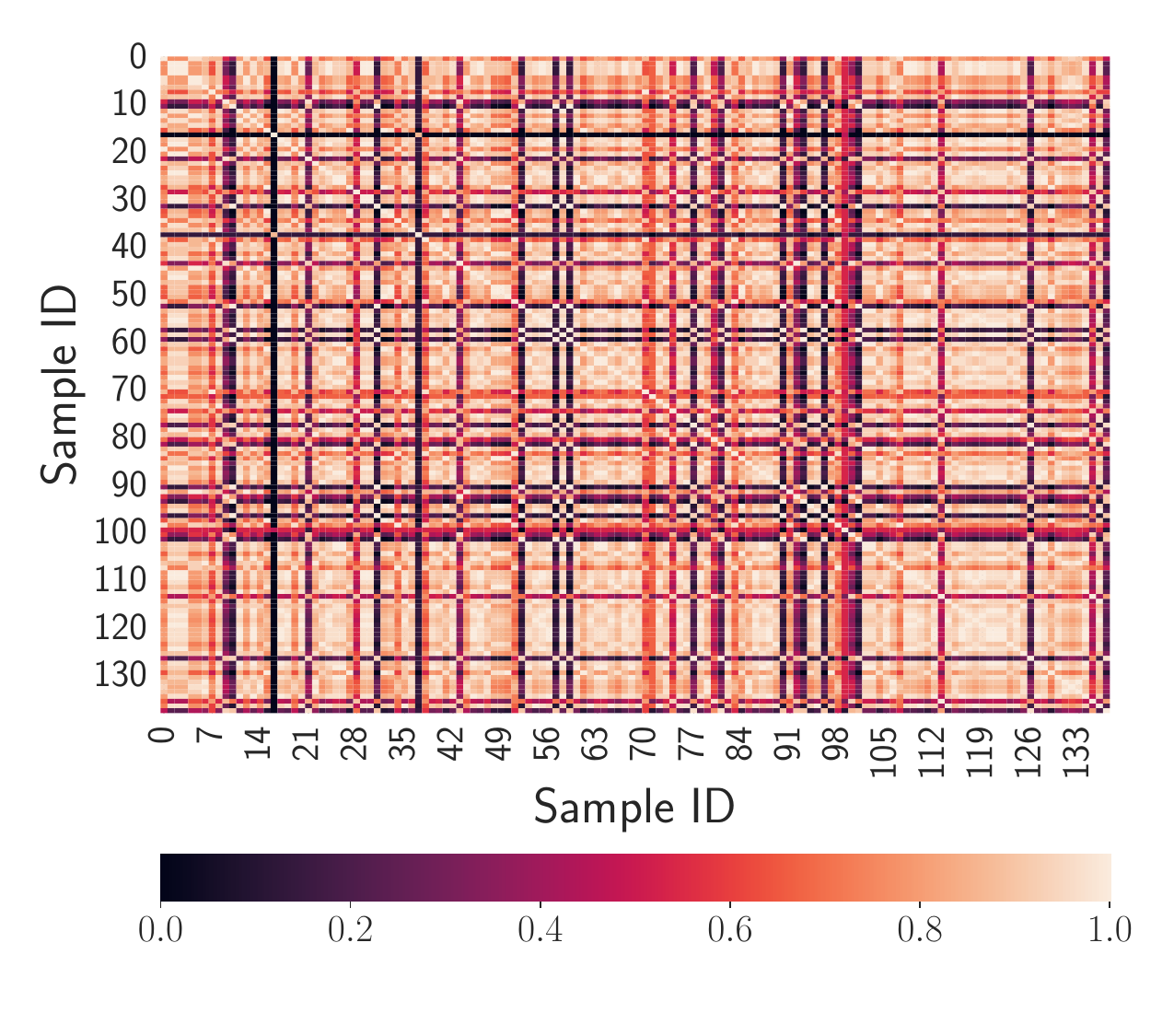}
    \end{subfigure}
    \begin{subfigure}{0.329\textwidth}
        \centering
        \includegraphics[width=\textwidth]{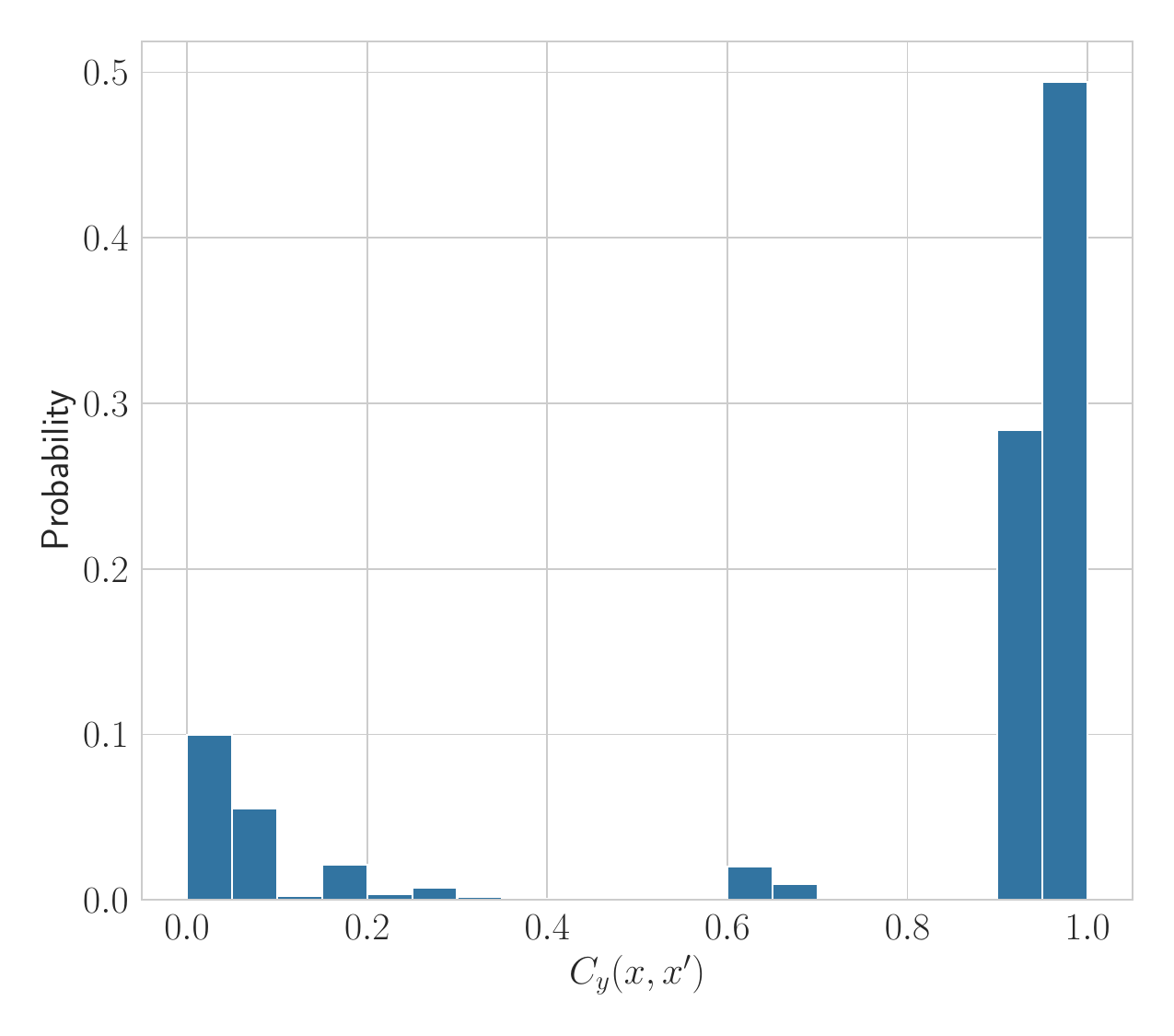}
        \caption{Person}
    \end{subfigure}
    \begin{subfigure}{0.329\textwidth}
        \centering
        \includegraphics[width=\textwidth]{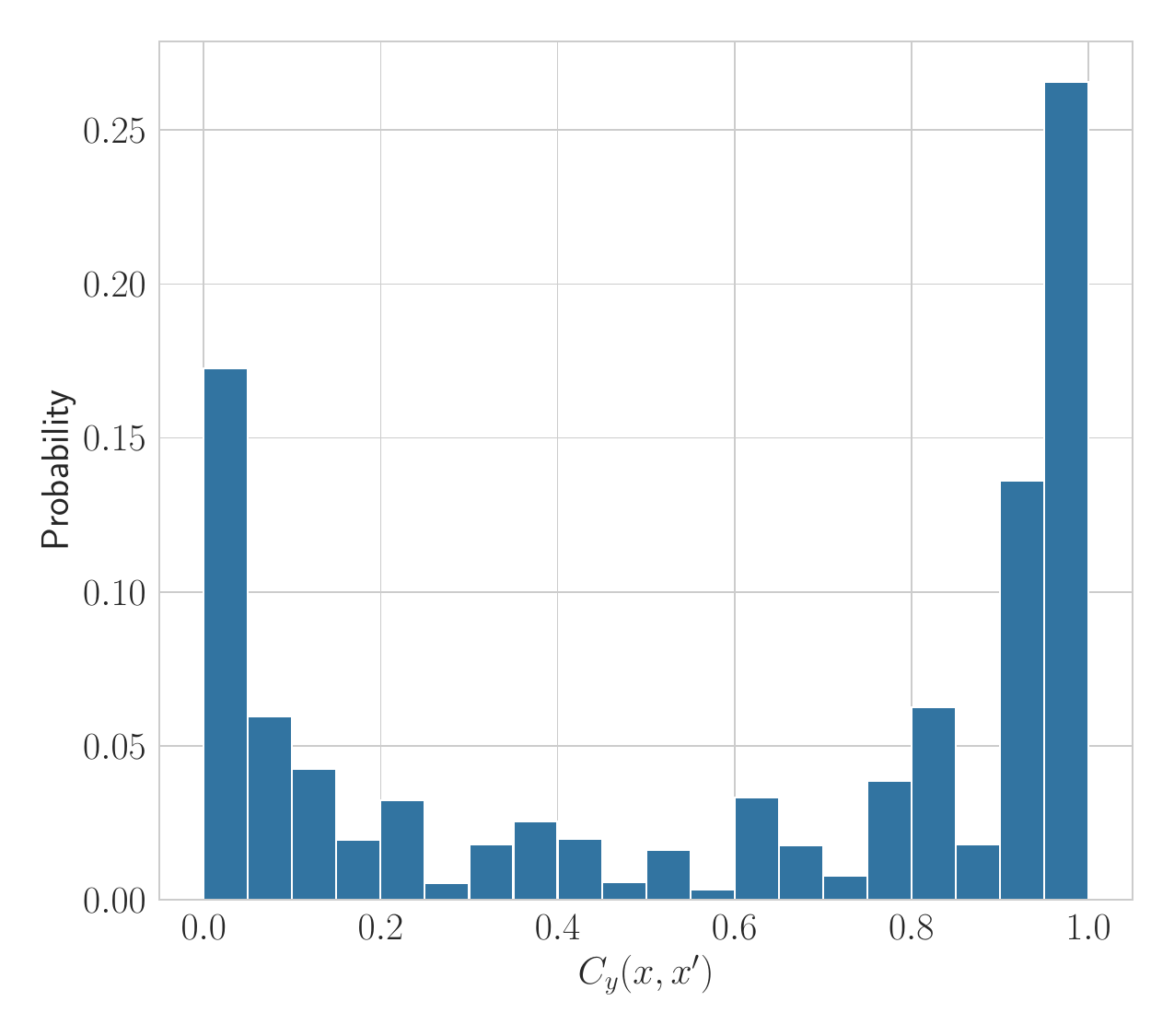}
        \caption{Entity}
    \end{subfigure}
    \begin{subfigure}{0.329\textwidth}
        \centering
        \includegraphics[width=\textwidth]{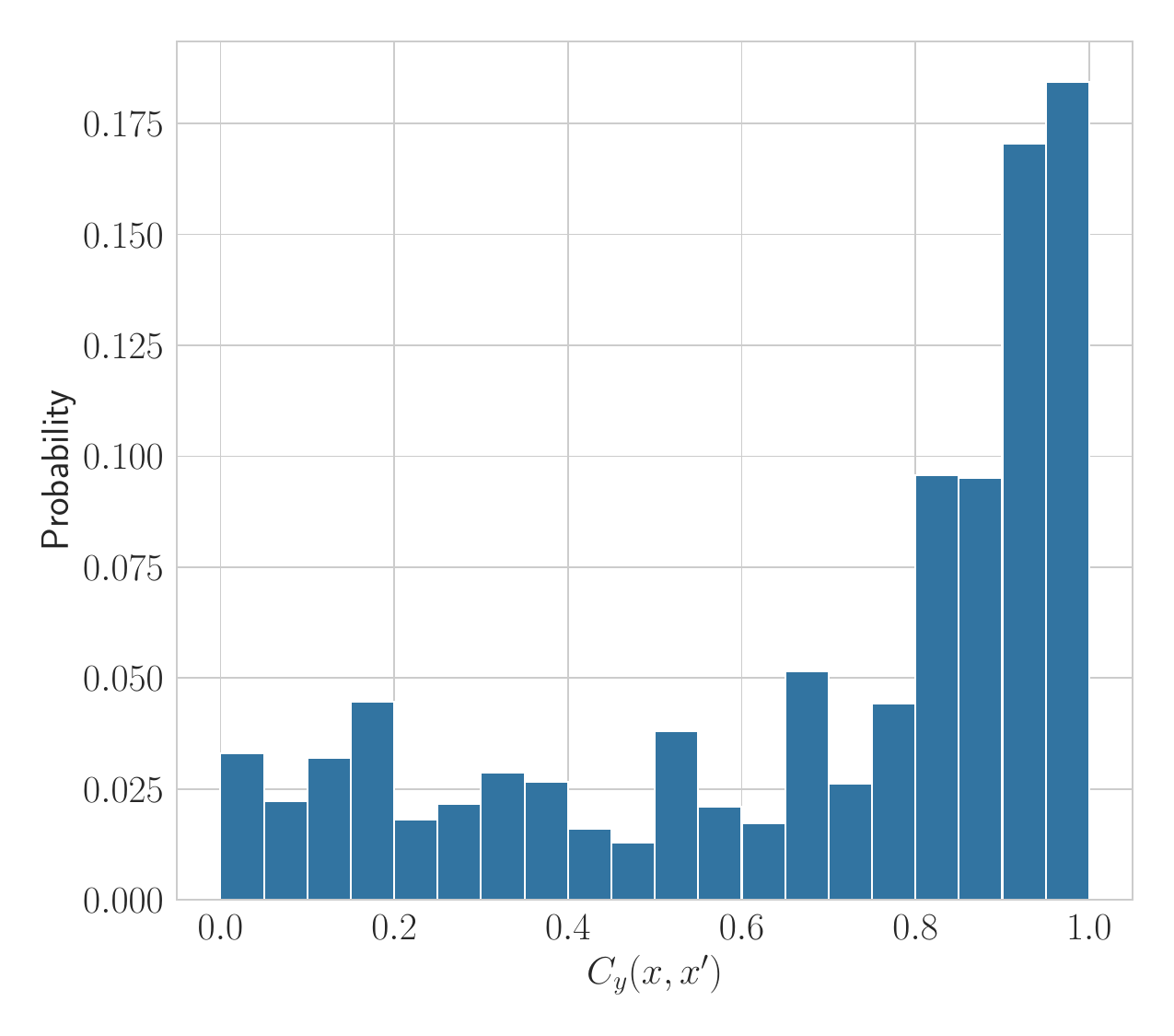}
        \caption{Description}
    \end{subfigure}
    
    \caption{\textit{Top:} we visualize the matrix of pairwise $C_y(\bm{x},\bm{x}')$ for three different TREC classes, using Llama3 as a classifier. \textit{Bottom:} we build a histogram for each of the above matrices, to show the distribution of consistency across samples of a given class.}
    \label{fig:consistency-analysis}
\end{figure*}
As a reminder, the pair-wise consistency tells us how much the distribution $p_\tau(\cdot|\bm{x}), \bm{x} \in \mathcal{D}_y$ differs from $p_\tau(\cdot|\bm{x}'), \bm{x}' \in \mathcal{D}_y$. Figure \ref{fig:consistency-analysis} (top) provides the matrix $C_y(\bm{x},\bm{x}')$ for three TREC classes, namely Person, Entity, and Description. Regarding the first class, we consistently observe high consistency except a few cases. By direct observation of the troublesome samples, we can devise prompting strategies targeted to them. In the case of the Entity class, the number of inconsistent pairs is higher than the number of consistent ones, whereby a batch of samples with IDs 46-50 are very consistent with each other but wholly inconsistent against all other samples, e.g., those with IDs 51-58. The former belong to different subclassees (\textit{color} and \textit{other}) than the latter (mostly \textit{animal}), hence defining subclasses in the prompt might provide a better semantic definition of the the class itself (we show how to do this in Appendix \ref{sec:consistency-improve-prompts}). Finally, the matrix of class Description has mixed values, with most of the probability mass being assigned to consistency values smaller than 0.75. Figure \ref{fig:consistency-analysis} (bottom) provides the histogram of these values that convey an aggregated view of these matrices. Compared to Figure \ref{fig:example-swap}, it is likely that effort spent improving consistency of the Entity class will also resolve the inconsistencies for the Description class, as these two are often confused by the LLM.

\begin{figure*}[t]
    \centering
    \begin{subfigure}{0.329\textwidth}
        \centering
        \includegraphics[width=\textwidth]{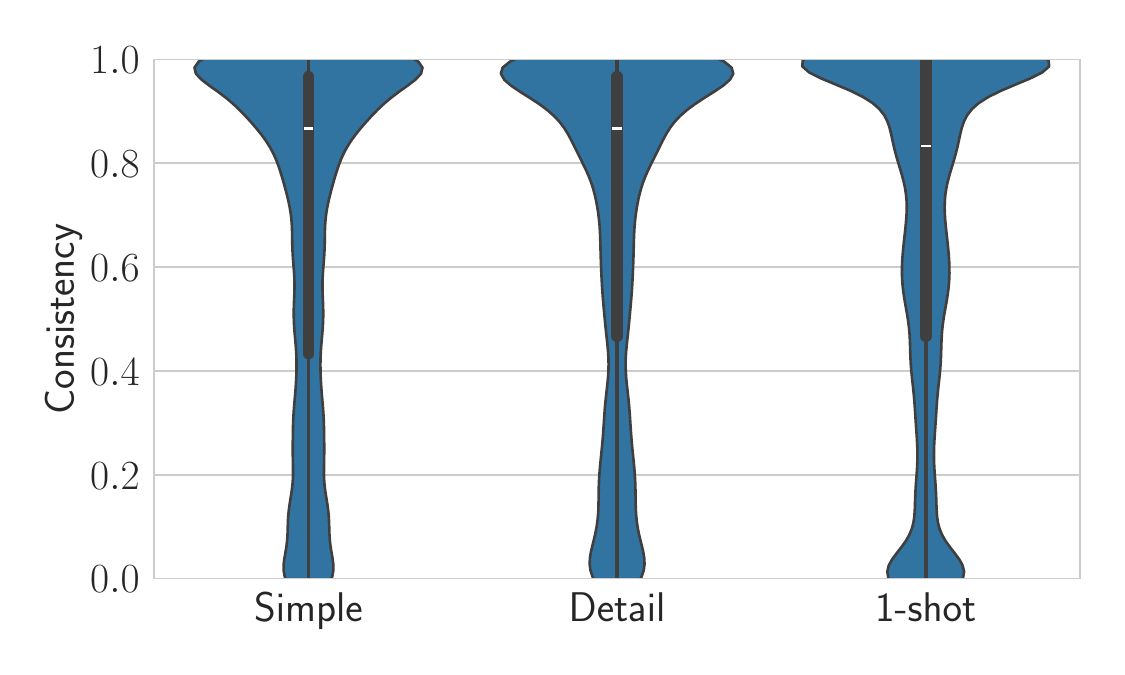}
        \caption{TREC}
    \end{subfigure}
    \begin{subfigure}{0.329\textwidth}
        \centering
        \includegraphics[width=\textwidth]{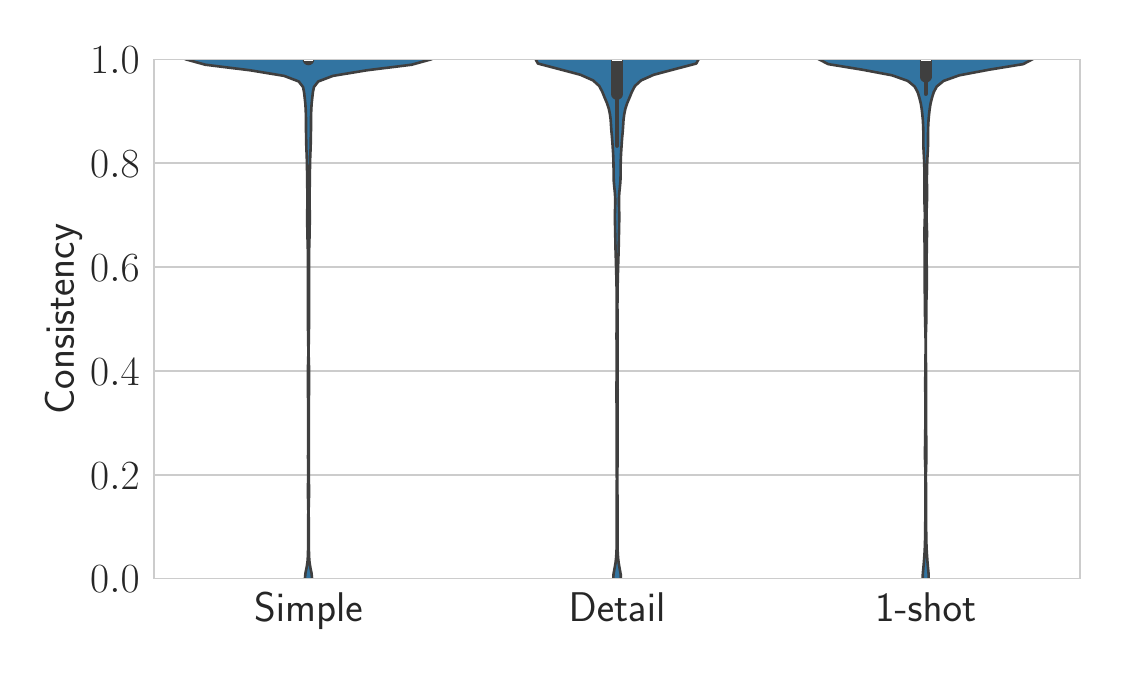}
        \caption{DBPedia}
    \end{subfigure}
    \begin{subfigure}{0.329\textwidth}
        \centering
        \includegraphics[width=\textwidth]{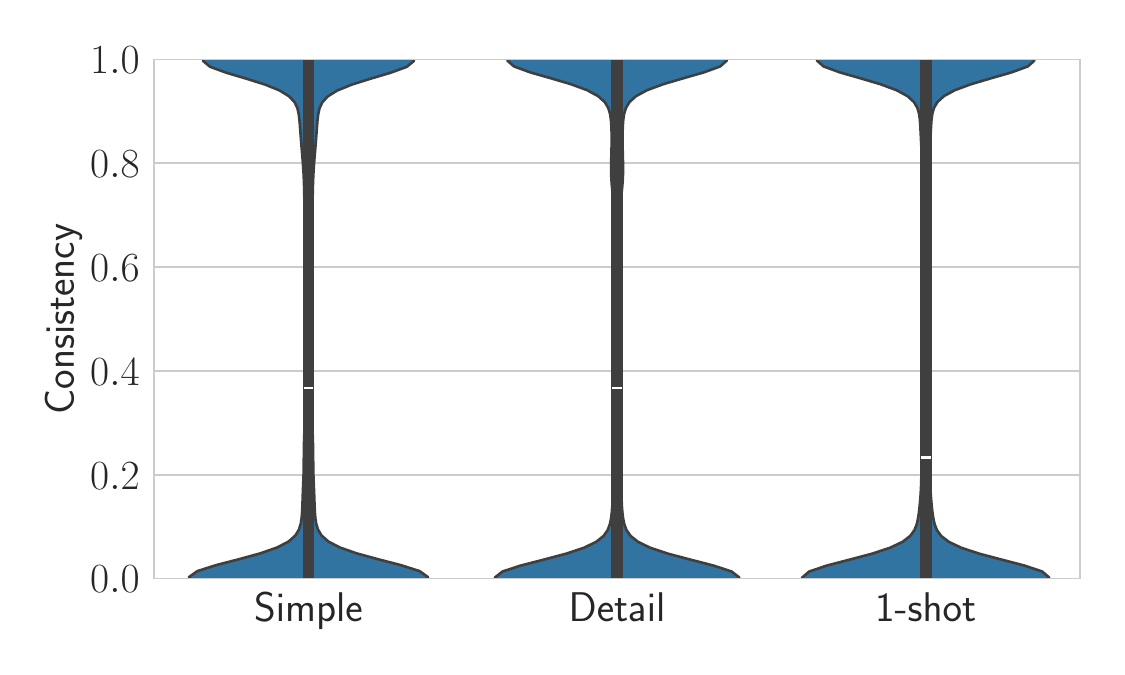}
        \caption{WoS}
    \end{subfigure}
    \caption{We show the violin plot of the Llama3 consistency over samples of the same classes, arranged by prompting technique, on different datasets.}
    \label{fig:consistency-violin}
\end{figure*}

Finally, we analyze the distribution of consistency values across the different prompting strategies, which is shown in Figure \ref{fig:consistency-violin}. There is an interesting pattern on TREC: the 1-shot strategy reduces the medium-level consistencies, but that does not necessarily imply an increase in all samples' consistency. There are more pairs of values with consistency 0 compared to not using a Simple or Detail strategy. These results show that qualitative results are more helpful than mean values: the behavior of the LLM is counter-intuitive compared to what one would expect, e.g., that a prompting strategy as the few-shot can only increase the consistency. Care should be put when analyzing such behaviors, and the prompt should be adapted accordingly. As regards the CB dataset, the distributions look very similar except for Detail: apparently, providing a class clarification makes the model more sensitive about its predictions for some classes, and we have shown earlier how to debug such faulty behavior.

\vspace{0.2em}
\take{Consistency finds sample groups misclassified similarly. Tuning prompts to large groups offers cost-benefit trade-offs.}

\section{Conclusions}
\label{sec:conclusions}
In this work, we have expanded the LLM developer's toolbox with two diagnostic metrics, namely sensitivity and consistency, that complement performance metrics such as accuracy. In our experiment, we showed how different prompting strategies influence these metrics and how we can guide prompt engineering decisions based on new criteria that value intrinsic characteristics of LLMs' predictions. Indeed, an LLM that is very sensitive to variation of the prompt and has high test set accuracy might not be a good choice in a production environment, where multiple intermediate steps exist and each could lead to minor alterations in the prompt of the LLM predictor. Notably, sensitivity does not require access to the ground truth labels, and it would be interesting in future work to extend it to tasks different from classification, for instance, code generation. Also, while this work mentioned prompt optimization, it is still an open question how to integrate these metrics in an automatic prompt engineering framework, leading to LLMs insensitive to nonsensical prompt variations and consistent in their (good) performances. Our hope is that sensitivity and consistency to the input will become relevant metrics from both an academic and industrial perspective, helping to identify pain points of LLMs.

\section{Limitations}
\label{sec:limitations}
The first clear limitation of the proposed metrics is that they work for classification problems only; despite classification being a common task in information extraction, extending at least sensitivity to more general problems is an important future work. Another inherent limitation is the trade-off between the quality of the approximation used to compute sensitivity and consistency, due to the (possibly biased) sampler $\mathcal{S}$, the number $Q$ of different prompt rephrasings, and the cost to run the extra queries. Future work should investigate if higher moments, such as the variance, of the metrics we have proposed provide more information without having access to class labels.

\section{Ethical Considerations}
\label{sec:ethical}
Our work proposes to evaluate LLMs according to metrics that gauge how well they perform when varying the input prompt and identify failure modes that need solving. For instance, we could use these metrics to discover that an LLM is more sensitive to a minority class than another, allowing us to solve the problem. Malicious attackers can use these metrics to understand if one LLM is more subject to jailbreaks than another, but they give no indication of how to do so. At the same time, making LLMs robust and optimized for these metrics may increase their trustworthiness.

\bibliography{bibliography}

\appendix
\clearpage

\section{Comparison with Noisy and Random Predictors}
\label{sec:comaprison-noise-random-predictors}
\begin{figure}[h]
    \centering
    \includegraphics[width=1\linewidth]{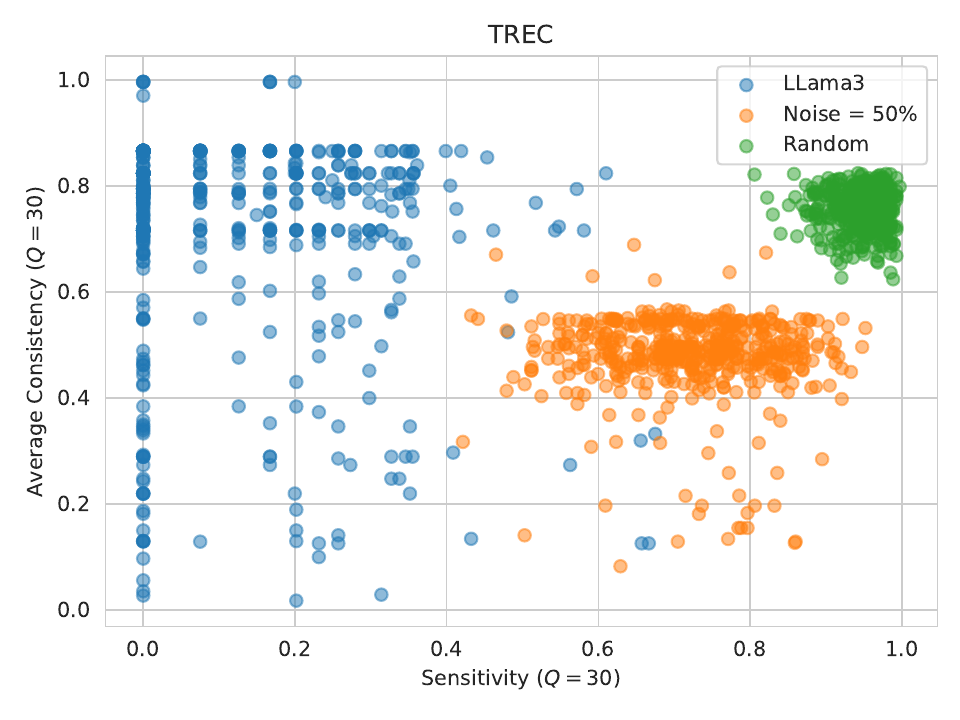}
    \caption{We plot, for each sample in TREC, its sensitivity and average consistency against all other samples.}
    \label{fig:noisy-random}
\end{figure}

To provide further evidence that $Q=30$ provides robust numbers that support our claims, we perturb the LLM predictions and observe that the joint metrics significantly deviate from the ones we report in Table \ref{tab:quantitative-results}. In particular, we consider a noisy predictor, where LLM's class predictions are randomly swapped 50\% of the time, and a completely random predictor. Results are shown in Figure \ref{fig:noisy-random} and based on LLama3 on TREC. We report, for each sample, its sensitivity and the average consistency against all other samples of the same class. These results clearly show that our choice of $Q=30$ produces a much different distribution of points compared to the perturbed versions of the LLM. Note, however, that $Q=30$ is not high enough to distinguish random consistency values from LLama3 in the specific case of TREC, since sampling is not perfectly uniform. That said, if observed jointly with sensitivity, a clear picture emerges. We encourage this kind of joint analysis since sensitivity and consistency both rely on the distribution $p_\tau(\cdot)$.

\section{Using Sensitivity to Improve Prompts}
\label{sec:sensitivity-improve-prompts}

To show that sensitivity can identify problematic samples, we sort TREC samples by descending order of LLama3 sensitivity, the first 10 most sensitive examples look like this:
\begin{itemize}
\itemsep 0em
    \item (0.67) When did the Hindenberg crash ?
    \item (0.67) When is the summer solstice ?
    \item (0.66) When was Algeria colonized ?
    \item (0.66) When was the Boston tea party ?
    \item (0.61) When did John F. Kennedy get elected as President ?
    \item (0.58) When was the telephone invented ?
    \item (0.57) When was Hiroshima bombed ?
    \item (0.56) What is Susan B. Anthony 's birthday ?
    \item (0.55) When did Hawaii become a state ?
    \item (0.54) When is St. Patrick 's Day ?
\end{itemize}
where the number in parenthesis indicates sensitivity value. The list continues, but all these examples seem to belong to the class Number, and they specifically refer to a date. We then refined the simple prompt strategy by adding the following sentence: \textit{``Note that questions about dates count as elements of class Number.''}. We recomputed sensitivity and noticed that the sensitivity scores dropped to [0.08, 0.0, 0.20, 0.27, 0.54, 0.17, 0.36, 0.08, 0.20, 0.24], which predictions over $Q=30$ prompt rephrasings heavily shifting in favor of the correct class: the amount of correct predictions over 30 rephrasings increases to [29, 30, 27, 26, 15, 27, 24, 29, 26, 26] compared to [17, 13, 14, 18, 10, 20, 21, 11, 19, 9] without the extra sentence in the prompt. This example demonstrates that we can identify weak spots of an LLM (in this case questions about dates) and try to correct its behavior via prompt engineering. Similar considerations applied for elements of class Entity whose input questions pertained to colors.

\section{Using Consistency to Improve Prompts}
\label{sec:consistency-improve-prompts}
We perform an experiment similar to that of the previous section, this time exploiting the samples obtained through the consistency analysis of Section \ref{sec:results}, which are different from the ones identified by sensitivity. In particular, examples of the class Entity with ID 46-50 are as follows:
\begin{itemize}
\itemsep 0em
    \item What color is indigo ?
    \item What does a barometer measure ?
    \item What color is a giraffe 's tongue ?
    \item What are the two types of twins ?
    \item What color is yak milk ?
\end{itemize}
By inspection of consistency matrices, we found that LLama3 has a hard time classifying elements of class Entity that refer to colors (exceptions are the second and fourth samples that belong to subclass ``other''). We hypothesize that all questions pertaining to colors might be problematic for LLama3, hence we refine the prompt by adding \textit{``Note that questions about colors count as elements of class Entity.''}. Once more, this prompt correction is enough to significantly improve the classification of the three examples about color, moving from a number of correct predictions over 30 rephrasings of [0, 4, 3, 15, 0] to [30, 4, 30, 15, 30].

Inconsistency of high-sensitivity samples is a good proxy for their ``classification hardness'', because their individual distributions $p_\tau$ differ a lot from the others. It is worth nothing that looking at standard misclassifications, which assume $Q=1$, is not enough to discover the same examples suggested by consistency; misclassification can only highlight if examples where wrongly classified. However, when samples of a given class are associated with non-negligible sensitivity (which requires higher $Q$ to be measured), then they can be recognized as very inconsistent compared to the other samples of the same class, and this effect is to be attributed mostly to the prompt by definition of sensitivity. In contrast, a quasi-zero sensitivity example that is (on average) inconsistent as well as misclassified suggests that the cause of the error is the intrinsic diversity of that sample from the others of the same class (see also the discussion in Section \ref{sec:method}). To empirically confirm our statements, we reuse LLama3 and sort samples of TREC class Description with lowest average consistencies, and filter them by non-zero sensitivity. The results are:
\begin{itemize}
\itemsep 0em
    \item (0.20) (0.13) What does PhiBetaKappa mean?
    \item (0.13) (0.35) What is the chunnel ?
    \item (0.13) (0.35) What is naproxen ?
    \item (0.13) (0.35) What is angiotensin ?
    \item (0.17) (0.37) What is e-coli ?
    \item (0.20) (0.40) What is amoxicillin ?
    \item (0.23) (0.43) What is Teflon ?
    \item (0.26) (0.45) What is acetaminophen ?
    \item (0.26) (0.45) What is semolina ?
    \item (0.36) (0.47) What does ciao mean ?
\end{itemize}
where numbers in brackets denote sensitivity and consistency, respectively. In this case, the number of correct predictions over $Q=30$ rephrasings of the original prompt is [0, 3, 3, 2, 5, 5, 4, 8, 8, 16]. If we enrich the prompt by adding \textit{``Note that questions starting with "What is" or "What are", or asking for the meaning of something generally refer to class Description.''}, these numbers increase to [22, 30, 30, 30, 29, 30, 30, 30, 30, 29], with sensitivity values dropping to 0 in 70\% of the cases and heavily decreasing in 90\% of cases. This shows that highly inconsistent samples (with non-zero sensitivity) allow us to further improve the prompts and correct mistakes that the LLM is bound to make when rephrasing the prompt. This can significantly enhance the LLMs' trustworthiness by guiding the creation of a comprehensive prompt. 

\section{Analysis of Standard Deviation Values}
\label{sec:std-results}
For completeness, we also report standard deviation values of our results. Please notice, as we write in Section \ref{sec:experiments}, that the distribution of results is not Gaussian and there is no reason it should be; therefore, statistics such as the standard deviation can convey misleading information. This is why we decided not to provide these numbers in the main paper and rather focus more on qualitative investigations, that analyze the actual distributions of sensitivity and consistency values and allow for a more meaningful analysis of failure cases. 

\begin{table*}[ht]
\small
\begin{tabular}{lcccccc}
\toprule
                     & \multicolumn{3}{c}{Llama3}                                                                                                                                                                                                       & \multicolumn{3}{c}{Mixtral}                                                                                                                                                                                                      \\ \midrule
\multicolumn{1}{c}{} & \begin{tabular}[c]{@{}c@{}}Simple\\ $S_\tau$/$C_y$/F1\end{tabular} & \begin{tabular}[c]{@{}c@{}}Detail\\ $S_\tau$/$C_y$/F1\end{tabular} & \multicolumn{1}{c}{\begin{tabular}[c]{@{}c@{}}1-shot\\ $S_\tau$/$C_y$/F1\end{tabular}} & \begin{tabular}[c]{@{}c@{}}Simple\\ $S_\tau$/$C_y$/F1\end{tabular} & \begin{tabular}[c]{@{}c@{}}Detail\\ $S_\tau$/$C_y$/F1\end{tabular} & \multicolumn{1}{c}{\begin{tabular}[c]{@{}c@{}}1-shot\\ $S_\tau$/$C_y$/F1\end{tabular}} \\ \midrule
TREC    & .148/.339/.085 & .149/.339/.080 & .146/.351/.051 & .168/.305/.071 & .182/.316/.056 & .158/.341/.050 \\
CB      & .082/.163/.005 & .073/.156/.004 & .067/.170/.006 & .216/.351/.048 & .188/.367/.049 & .188/.360/.076 \\
RTE     & .244/.397/.066 & .289/.389/.121 & .160/.402/.013 & .277/.376/.037 & .263/.383/.041 & .152/.401/.010  \\
DBPedia & .060/.202/.011 & .095/.217/.062 & .076/.210/.031 & .104/.321/.025 & .113/.332/.014 & .115/.329/.020 \\
WoS     & .103/.466/.008 & .091/.472/.008 & .092/.472/.008 & .164/.424/.019 & .171/.416/.013 & .171/.401/.028 \\ \bottomrule
\end{tabular}
\begin{tabular}{lcccccc}
\toprule
                     & \multicolumn{3}{c}{GPT-3.5}                                                                                                                                                                                                       & \multicolumn{3}{c}{GPT-4o}                                                                                                                                                                                                      \\ \midrule
\multicolumn{1}{c}{} & \begin{tabular}[c]{@{}c@{}}Simple\\ $S_\tau$/$C_y$/F1\end{tabular} & \begin{tabular}[c]{@{}c@{}}Detail\\ $S_\tau$/$C_y$/F1\end{tabular} & \multicolumn{1}{c}{\begin{tabular}[c]{@{}c@{}}1-shot\\ $S_\tau$/$C_y$/F1\end{tabular}} & \begin{tabular}[c]{@{}c@{}}Simple\\ $S_\tau$/$C_y$/F1\end{tabular} & \begin{tabular}[c]{@{}c@{}}Detail\\ $S_\tau$/$C_y$/F1\end{tabular} & \multicolumn{1}{c}{\begin{tabular}[c]{@{}c@{}}1-shot\\ $S_\tau$/$C_y$/F1\end{tabular}} \\ \midrule
TREC    & .174/.387/.026 & .164/.371/.034 & .155/.385/.018 & .099/.344/.013 & .120/.367/.030 & .096/.335/.010 \\
CB      & .277/.284/.028 & .265/.252/.029 & .252/.260/.025 & .174/.313/.012 & .164/.306/.013 & .145/.266/.014 \\
RTE     & .229/.267/.023 & .232/.245/.025 & .234/.309/.015 & .295/.300/.053 & .187/.305/.006 & .168/.302/.006 \\
DBPedia & .071/.230/.007 & .081/.258/.012 & .070/.228/.008 & .069/.201/.015 & .068/.229/.015 & .057/.230/.007 \\
WoS     & .140/.430/.018 & .138/.431/.018 & .141/.421/.035 & .105/.461/.004 & .100/.463/.003 & .100/.463/.003 \\ \bottomrule
\end{tabular}
\caption{Standard deviation values of sensitivity, consistency, and micro-F1 score across all prompt rephrasings $\rho$ are shown for different datasets, models, and prompting strategies. We remind the reader that standard deviation values can be misleading when the distribution is not Gaussian (which is the case in our experiments).}
\end{table*}

\section{Examples of Prompt Rephrasings}
\label{sec:prompt-rephrasing}

We report below some task rephrasings that we generated for the different datasets using LLama3. The complete list of rephrasings for the other models is similar and can be inspected in the supplementary material. Please note that these sentences are just a part of the whole prompt and they are used to describe the task; the rest of the prompt remains the same for all models. We did not observe strange behavior in the rephrasings produced by the LLM. The first sentence corresponds to the original prompt.

\paragraph{TREC}

\begin{itemize}
\itemsep0em 

\item "Classify the questions based on whether their answer type is a Number, Location, Person, Description, Entity, or Abbreviation."
\item "What categories - Number, Location, Person, Description, Entity, or Abbreviation - do the answers to the questions fall under?"
\item "Into which of the following categories - Number, Location, Person, Description, Entity, or Abbreviation - do the answers to the questions belong?"
\item "How would you categorize the answers to the questions: are they numerical values, specific locations, individuals, descriptive phrases, entities, or abbreviated forms?"
\item "Can the answers to the questions be grouped into one of the six categories: numerical values, geographic locations, individuals, descriptive phrases, entities, or abbreviated forms?"
\item "What classification system, comprising Number, Location, Person, Description, Entity, and Abbreviation, can be applied to the answers of the questions?"
\item "Organize the questions according to the type of answer they require, which can be one of the following: a numerical value, a specific place, a person's name, a descriptive phrase, an entity, or an abbreviated term."
\item "Based on their answer types, sort the questions into the categories of numeric responses, geographic locations, personal names, descriptive phrases, entities, or abbreviated terms."
\item "What is the appropriate categorization for the answers to the questions, whether they represent a quantitative value, a geographic point, an individual, a descriptive passage, a distinct entity, or a shortened form?"
\item "Categorize the questions by the type of information their answers provide, which may be a numerical value, a specific location, a person's identity, a descriptive explanation, a distinct entity, or an abbreviated term."
\end{itemize}

\paragraph{CB}

\begin{itemize}
\itemsep0em 

\item "You are given a premise and a hypothesis as input. Determine is there is entailment, contradiction, or a neutral relation between the premise and the hypothesis."
\item "What is the logical relationship between a given premise and hypothesis: do they entail each other, contradict, or have no logical connection?"
\item "Classify the relationship between a premise and a hypothesis as either entailment, contradiction, or neutrality."
\item "Can a given premise be said to logically imply, contradict, or have no bearing on a corresponding hypothesis?"
\item "Identify the nature of the relationship between a premise and a hypothesis, specifically whether it represents a logical implication, a contradiction, or a lack of connection."
\item "What kind of relationship exists between a premise and a hypothesis: do they logically align, clash, or remain unrelated?"
\item "Determine whether the premise and hypothesis, when considered together, yield a logical inference, a contradictory stance, or a lack of logical correlation."
\item "Is the hypothesis supported, refuted, or unrelated to the premise, based on their logical connection?"
\item "Examine the logical bond between a premise and a corresponding hypothesis to ascertain whether they are logically compatible, mutually exclusive, or independent of each other."
\item "What is the nature of the logical bond between a premise and a hypothesis: does it represent a logical implication, a contradictory stance, or a lack of logical correlation?
\end{itemize}

\paragraph{RTE}

\begin{itemize}
\itemsep0em 

\item "You are given a premise and a hypothesis as input. Determine is there is entailment or a contradiction/neutral statement."
\item "Can you identify whether the given premise logically implies or contradicts the provided hypothesis, or if they are unrelated?"
\item "Do the input premise and hypothesis have a logical relationship, specifically entailment, contradiction, or neutrality?"
\item "What is the logical relationship between the input premise and hypothesis: does the premise imply, contradict, or neither imply nor contradict the hypothesis?"
\item "Is the input hypothesis supported, refuted, or unrelated to the given premise based on logical inference?"
\item "Classify the relationship between the input premise and hypothesis as one of entailment, contradiction, or neutrality."
\item "Can the input premise be logically connected to the hypothesis in a way that implies, refutes, or has no bearing on the hypothesis?"
\item "Determine whether the input premise logically supports, contradicts, or is independent of the given hypothesis."
\item "What can be inferred about the relationship between the premise and hypothesis: do they logically align, contradict each other, or remain unrelated?"
\item "Is the input hypothesis a logical consequence of the given premise, or does it contradict or have no logical connection to the premise?"
\end{itemize}
\paragraph{DBPedia}

\begin{itemize}
\itemsep0em 

\item "Classify the text based on whether their subject is a Company, Educational Institution, Artist, Athlete, Office Holder, Mean Of Transportation, Building, Natural Place, Village, Animal, Plant, Album, Film, or Written Work."
\item "What category does the subject of the text belong to: Company, Educational Institution, Artist, Athlete, Office Holder, Mean Of Transportation, Building, Natural Place, Village, Animal, Plant, Album, Film, or Written Work?"
\item "Into which of the following categories does the subject of the text fall: Company, Educational Institution, Artist, Athlete, Office Holder, Mean Of Transportation, Building, Natural Place, Village, Animal, Plant, Album, Film, or Written Work?"
\item "Is the subject of the text a type of organization, such as a Company or Educational Institution, a person, like an Artist, Athlete, or Office Holder, a mode of transportation, a structure, a location, a living thing, or a creative work?"
\item "Categorize the text according to the type of entity its subject represents, choosing from the following options: Company, Educational Institution, Artist, Athlete, Office Holder, Mean Of Transportation, Building, Natural Place, Village, Animal, Plant, Album, Film, or Written Work."
\item "What type of entity is the subject of the text: a corporate entity, an educational establishment, a creative individual, a sports figure, a government official, a vehicle, a constructed facility, a geographical location, a small settlement, a living creature, a botanical organism, a music collection, a motion picture, or a literary composition?"
\item "To which of the following categories does the subject matter of the text correspond: Company, Educational Institution, Artist, Athlete, Office Holder, Mean Of Transportation, Building, Natural Place, Village, Animal, Plant, Album, Film, or Written Work?"
\item "Identify the category that best describes the subject of the text, selecting from the options of Company, Educational Institution, Artist, Athlete, Office Holder, Mean Of Transportation, Building, Natural Place, Village, Animal, Plant, Album, Film, or Written Work."
\item "Determine the classification of the text's subject, which can be one of the following: a business organization, a school or university, a creative person, a sports personality, a government position, a vehicle, a constructed structure, a geographical location, a small town, a creature, a botanical species, a music release, a movie, or a written piece."
\item "What is the primary topic of the text: a corporate entity, a place of learning, a creative individual, a sports figure, a government official, a vehicle, a man-made structure, a natural location, a small community, a living organism, a botanical species, a music collection, a motion picture, or a written composition?"
\end{itemize}

\paragraph{WoS}

\begin{itemize}
\itemsep0em 

\item "Classify the text based on whether their field is Computer Science, Electrical Engineering, Psychology, Mechanical Engineering, Civil Engineering, Medical Science, or Biochemistry."
\item "What category does the text belong to: Computer Science, Electrical Engineering, Psychology, Mechanical Engineering, Civil Engineering, Medical Science, or Biochemistry, based on the field it represents?"
\item "Into which of the following fields does the text fall: Computer Science, Electrical Engineering, Psychology, Mechanical Engineering, Civil Engineering, Medical Science, or Biochemistry?"
\item "Categorize the text according to the field it pertains to, choosing from the options of Computer Science, Electrical Engineering, Psychology, Mechanical Engineering, Civil Engineering, Medical Science, or Biochemistry."
\item "Identify the field of study represented in the text, selecting from the options of Computer Science, Electrical Engineering, Psychology, Mechanical Engineering, Civil Engineering, Medical Science, or Biochemistry."
\item "Determine the discipline that the text corresponds to, selecting from among Computer Science, Electrical Engineering, Psychology, Mechanical Engineering, Civil Engineering, Medical Science, and Biochemistry."
\item "Which of the seven fields - Computer Science, Electrical Engineering, Psychology, Mechanical Engineering, Civil Engineering, Medical Science, or Biochemistry - does the text's subject matter align with?"
\item "What is the academic discipline that the text is related to, with possibilities including Computer Science, Electrical Engineering, Psychology, Mechanical Engineering, Civil Engineering, Medical Science, or Biochemistry?"
\item "Assign a category to the text from the following options: Computer Science, Electrical Engineering, Psychology, Mechanical Engineering, Civil Engineering, Medical Science, or Biochemistry, based on the field of study it describes."
\item "Based on the field of study, sort the text into one of the following categories: Computer Science, Electrical Engineering, Psychology, Mechanical Engineering, Civil Engineering, Medical Science, or Biochemistry."
\end{itemize}

\end{document}